\documentclass[preprint,authoryear,12pt]{elsarticle}

\usepackage{graphics}

\usepackage{amssymb}
\usepackage{amsthm}

\usepackage{array}
\usepackage{amsmath}
\usepackage{booktabs}
\usepackage{url}
\usepackage{epsfig}
\usepackage{epstopdf}
\usepackage{float}
\usepackage[caption=false,font=footnotesize]{subfig}
\usepackage{multirow}
\newcommand{\tabincell}[2]{\begin{tabular}{@{}#1@{}}#2\end{tabular}}  

\begin{document}

\begin{frontmatter}



\title{HEp-2 Cell Image Classification with Deep Convolutional Neural Networks}


\author[label1]{Zhimin~Gao}
\ead{zg126@uowmail.edu.au}
\cortext[cor1]{Corresponding author}
\author[label1]{Lei~Wang\corref{cor1}}
\ead{leiw@uow.edu.au}
\author[label1]{Luping~Zhou}
\ead{lupingz@uow.edu.au}
\author[label1]{Jianjia~Zhang}
\ead{jz163@uowmail.edu.au}

\address[label1]{School of Computer Science and Software Engineering, University of Wollongong, NSW 2522, Australia}

\begin{abstract}
Efficient Human Epithelial-2 (HEp-2) cell image classification can facilitate the diagnosis of many autoimmune diseases. This paper presents an automatic framework for this classification task, by utilizing the deep convolutional neural networks (CNNs) which have recently attracted intensive attention in visual recognition. This paper elaborates the important components of this framework, discusses multiple key factors that impact the efficiency of training a deep CNN, and systematically compares this framework with the well-established image classification models in the literature. Experiments on benchmark datasets show that i) the proposed framework can effectively outperform existing models by properly applying data augmentation; ii) our CNN-based framework demonstrates excellent adaptability across different datasets, which is highly desirable for classification under varying laboratory settings. Our system is ranked high in the cell image classification competition hosted by ICPR 2014.  
\end{abstract}


\end{frontmatter}


\section{Introduction}
\label{sec:introduction}
Indirect immunofluorescence (IIF) on Human Epithelial-$2$ (HEp-$2$) cells is a recommended methodology to diagnose autoimmune diseases \citep{rigon2007indirect}. However, manual analysis of IIF images leads to crucial limitations, such as the subjectivity of result, the inconsistence across laboratories, and the low efficiency in processing a large number of cell images \citep{meroni2010ana, Benchmarking2012}. To improve this situation, automatic and reliable cell images classification has become an active research topic. 

Many methods have been recently proposed for this topic, especially during the HEp-$2$ cell classification competitions \citep{Benchmarking2012, Foggia20142305, ICPR2014Report}. Most of them treat feature extraction and classification as two separate stages. For the former, a variety of hand-crafted features are adopted, including local binary pattern (LBP) \citep{lbp1990, Nosaka20142428, Theodorakopoulos20142367}, scale-invariant feature transform (SIFT) \citep{lowe2004distinctive}, histogram of oriented gradients \citep{hog2005}, discrete cosine transform, and the statistical features like gray-level co-occurrence matrix \citep{glcm1973} and gray-level size zone matrix \citep{tbme14statiscal}. For the latter, nearest-neighbor classifier, boosting, support vector machines (SVM) and multiple kernel SVM have been employed \citep{Wiliem20142315}. As a result, the performance of these classifiers relies highly on the appropriateness of the empirically chosen hand-crafted features. Moreover, because features and classifier are treated separately, they cannot work together to maximally identify and retain discriminative information.

Very recently, deep convolutional neural networks (CNNs) have consistently achieved outstanding performance on generic visual recognition tasks \citep{ImageNet12} and this has revived extensive research interest in CNN-based classification model \citep{cnnoff2014}. The CNNs consist of multi-stage processing of an input image to extract hierarchical and high-level feature representations. Many hand-crafted features and the corresponding classification pipelines can be regarded as an approximation to or a special case of the CNNs, by sharing some basic building blocks. Nevertheless, these features and pipelines have to be carefully designed and integrated in order to preserve discriminative information. The excellent performance achieved by deep CNNs on generic visual recognition and the high demand for full automation of HEp-$2$ cell image classification motivate us to research the CNNs for this classification task.
   
To this end, we propose an automatic feature extraction and classification framework for HEp-$2$ staining patterns based on deep CNNs \citep{IEEELeCun98}. This framework extracts features from the raw pixels of cell images and avoids using hand-crafted features. Feature representations for each kind of staining patterns are learned and optimized via training the multi-layer network. Also, the classification layer is jointly learned with this network to predict the probability of a cell image for each class. The highly non-linear and high-capacity properties \citep{lecun2012efficient} make the multi-layer CNNs difficult to train, especially when the number of training samples is not sufficiently large. We explore multiple important aspects in this CNN-based classification system, including network architecture, image preprocessing, hyper-parameters selection, and data augmentation, which are important for CNNs to achieve effective and reliable cell classification. Furthermore, we conduct rigorous experimental comparison with two state-of-the-art hand-designed shallower image representation models, i.e., bag-of-features (BoF) and Fisher Vector (FV), to investigate the advantages and disadvantages of our CNN-based framework on cell image classification. Our system has participated in the \emph{Contest on Performance Evaluation on Indirect Immunofluorescence Image Analysis Systems} hosted by ICPR 2014\footnote{Contest website is at \url{http://i3a2014.unisa.it/?page_id=91.}} and won the fourth place among $11$ international teams. 

The rest of the paper is organized as follows. Section \ref{sec:related-work} reviews the classification models of BoF, FV and deep CNNs. In Section \ref{sec:proposed-framework}, our CNN-based framework for cell images classification is presented and a set of key factors are discussed. Section \ref{sec:experiment} reports the experimental investigation and comparison, and the conclusions are drawn in Section \ref{sec:conclusion}.

We were invited by the ICPR 2014 contest organizers to report our system in a workshop short paper \citep{icpr14our}. This paper significantly extends that workshop paper in the following aspects: i) a more detailed description of our deep CNN-based classification framework for HEp-$2$ cell images is presented and multiple key factors for effectively training a reliable deep CNN are discussed and experimentally demonstrated; ii) the role of image rotation as a data augmentation method in helping the deep CNN to achieve robust representations in this classification task is investigated and analyzed; iii) systematic experimental comparisons of our CNN-based framework and the state-of-the-art hand-designed classification models are conducted; iv) the excellent adaptability of our cell classification system with respect to different laboratory settings is demonstrated by transferring the learned network across two datasets with easy implementation, which makes our system attractive for practical clinical applications.

\section{Related Work} \label{sec:related-work}
\subsection{Bag-of-features and Fisher Vector Models} 
The BoF model \citep{csurka2004visual} generally consists of four stages: local feature extraction, dictionary learning, feature encoding, and feature pooling. The dictionary is composed of a set of visual words describing the common visual patterns shared by local descriptors. The relationship between local descriptors and visual words is characterized by feature encoding. A variety of coding methods have been proposed in the literature \citep{liu2011defense, wang2010locality, VLAD, boiman2008defense}. On top of these, spatial pyramid matching (SPM) \citep{lazebnik2006beyond} is usually utilized to incorporate the spatial information of an image. The BoF model has been applied to staining patterns classification \citep{Wiliem20142315, Kong20142379, Shen20142419, Stoklasa20142409}, in which one or more of the above four stages are tailored to obtain better cell image representations for classification. Readers are referred to the review \citet{Foggia20142305} for more details.

In the past several years, FV model has shown superior performance to the BoF model \citep{perronnin2007fisher, perronnin2010improving, sanchez2013image}. Their main differences lie at dictionary learning and feature encoding. The dictionary in FV is generated by a probabilistic model, e.g., the Gaussian mixture model (GMM), that characterizes the distribution of local descriptors. Each local descriptor is then encoded by the first- and second-order gradients with respect to the model parameters. FV model has also been applied to cell image classification \citep{Faraki20142348, Han6805614fisher}.
\subsection{Deep Convolutional Neural Networks} 
CNNs belong to a class of learning models inspired by the multi-stage processes of visual cortex \citep{hubel1962receptive}. A pioneering work of CNNs was Fukushima's ``neocognitron'' \citep{Fukushima}. It has a structure similar to the hierarchical model of the visual nervous system discovered by Hubel and Wiesel \citep{hubel1959receptive}. Each stage of the network imitates the functions of simple and complex cells in the primary visual cortex. Later on, \Citet{IEEELeCun98} extended the neocognitron by utilizing backpropagation algorithm to train the model parameters of CNNs and achieved excellent performance in hand-written digit recognition.

With the advent of fast parallel computing, better regularization strategies, and large-scale datasets, deep CNNs models have recently significantly outperformed the models with hand-crafted features on generic object classification, detection and retrieval \citep{cnnoff2014}, as well as other visual recognition tasks, such as face verification \citep{deepface2014} and mitosis detection in breast cancer histopathology images \citep{Veta2015237}. As for cell images classification, Malon et al. \citep{Benchmarking2012} adopted a CNN to classify HEp-$2$ cell images.  \Citet{buyssens2013multiscale} designed a multiscale CNN for cytological pleural cancer cells classification. Our CNN framework presented in this paper is different from their works in terms of both image preprocessing method and network architecture. Moreover, our CNN performs better than the CNN reported in \citet{Benchmarking2012} on ICPR 2012 HEp-$2$ cell classification. 

Although CNNs have been initially applied to cell image classification, the following issues have not been systematically investigated and thus remain unclear: i) what are the key issues when adopting deep CNNs for cells classification? ii) how is the performance of the CNN-based classification model when compared with the well-established classification models in the literature, especially the BoF and FV models? These issues will be carefully investigated and addressed in this work.

\section{Proposed Framework} \label{sec:proposed-framework}
The proposed deep CNN-based HEp-2 cell image classification framework consists of three components: image preprocessing, network training, and feature extraction and classification, which are elaborated in this section. Also, data augmentation which plays an important role in this classification framework will be described and analyzed.

\begin{figure}[h]
\centering
\includegraphics[width=13.6cm]{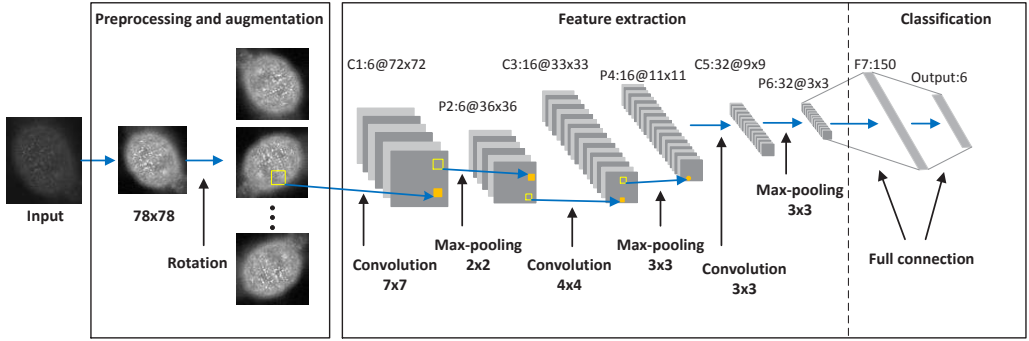}
\caption{The architecture of our deep convolutional neural network classification system for HEp-$2$ cell images. Each plane within the feature extraction stage denotes a feature map. The convolutional layer and max-pooling layer is abbreviated as C and P respectively. C1:6@$72\times72$ means that this is a convolutional layer, and is the first layer of the network. This layer is comprised of six feature maps, each of which has size of $72\times72$. The symbols and number above the feature maps of other layers have the similar meaning, whereas F7:150 means that this is a fully-connected layer. It is the seventh layer of the network and has $150$ neurons. The words and number between two layers stand for: the operation, i.e., convolution or max-pooling, applied to the feature maps of the previous layer in order to obtain the feature maps of this layer; and the size of each filter or the size of pooling region.}
\label{fig:architecture}
\end{figure}

\subsection{Network Architecture} \label{subsec:net-archi}
A proper selection of network architecture is crucial to CNNs. Usually, deep CNNs are composed of multiple convolutional layers interlaced with subsampling (pooling) layers, as shown in Fig. \ref{fig:architecture}. Each layer outputs a set of two-dimensional feature maps, each of which represents a specific feature detected from all positions of the input. These feature maps are in turn used as the input of the next layer. Fully-connected layers are usually stacked on the top of the network to conduct classification. 

Our deep CNN shares the basic architecture as the classical LeNet-5 \citep{IEEELeCun98}. Specifically, it contains eight layers. Among them, the first six layers are convolutional layers alternated with pooling layers, and the remaining two are fully-connected layers for classification. 

\subsubsection{Convolutional Layer} \label{subsubsec:conv-layer}
Let's assume that it is the $l$th layer. Let $N^l$ denote the number of feature maps at this layer, where $l$ is used as a superscript. Accordingly, each feature map is denoted as $\textbf{h}_j^{l}~(j = 1, 2, ..., N^l)$. This convolutional layer is parametrized by an array of two-dimensional filters $\textbf{W}_{ij}^l$ associating the $i$th feature map $\textbf{h}_i^{l-1}$ in the $(l-1)$th layer with the $j$th feature map $\textbf{h}_j^{l}$ in the $l$th layer and the bias $b_{j}$. Each filter acts as a feature detector to detect one particular kind of feature by convolving with every location of the input feature map. To obtain $\textbf{h}_j^{l}$, each input feature map $\textbf{h}_i^{l-1}~(i = 1, 2, ..., N^{l-1})$ is firstly convolved with the corresponding filter $\textbf{W}_{ij}^l$. The results are summed and appended with the bias $b_j^l$. After that, a non-linear activation function $\phi(\cdot)$, which can be sigmoid, tanh or rectified linear function \citep{ImageNet12}, is applied in an element-wise manner. Mathematically, the feature maps of the $l$th layer can be expressed as follows:
\begin{equation}
\textbf{h}_j^l = \phi(\sum_{i=1}^{N^{l-1}}\textbf{h}_i^{l-1} * \textbf{W}_{ij}^l + b_j^l),~j = 1, 2, ..., N^l.
\end{equation}
where $*$ denotes the convolution operation.

\subsubsection{Pooling Layer} \label{subsubsec:pooling-layer}
A pooling layer down-samples a feature map. This will greatly reduce the computation of training a CNN and also introduces invariance to small translations of input images. Max-pooling or average-pooling is usually applied. The former selects the maximum activation over a small pooling region, while the latter uses the average activation over this region. Max-pooling generally performs better than average-pooling \citep{boureau2010theoretical}. 

\subsubsection{Classification Layer} \label{subsubsec:class-layer}
Classification layers usually involve one or more fully-connected layers at the top of a CNN. Our network contains two fully-connected layers. The first fully-connected layer (F7 in Fig. \ref{fig:architecture}) takes the cascade of all the feature maps of the sixth layer (denoted as $\textbf{h}^{6}$) as input. This layer is parametrized by weights $\textbf{W}^7$ and biases $\textbf{b}^7$. The output of this layer $\textbf{h}^7$ is obtained as $\textbf{h}^7 =  \phi(\textbf{W}^7 \textbf{h}^{6} + \textbf{b}^7)$. The last fully-connected layer is the output layer and parametrized by weights $\textbf{W}^8$ and biases $\textbf{b}^8$. It contains $n$ neurons corresponding to $n$ classes of staining patterns, and outputs the probabilities ${\boldsymbol{\hat y}} = [\hat y_1, \hat y_2,..., \hat y_n]^\top \in \mathbb{R}^{n}$ via softmax regression as follows:
\begin{eqnarray}
\textbf{h}^8 &= &\textbf{W}^8 \textbf{h}^{7} + \textbf{b}^8,~\textbf{h}^8\in \mathbb{R}^{n}\\
\hat y_j &= &\frac{\exp(h_j^8)}{\sum_{i=1}^{n}\exp(h_i^8)},~j = 1, 2, ..., n.
\end{eqnarray}
where $\hat y_j $ is the output probability of the $j$th neuron.

The network architecture of our deep CNN is illustrated in Fig. \ref{fig:architecture}. Specifically, the first layer convolves an input image with each of the six filters of size $7 \times 7$ with a stride of one pixel, and then adds a bias to each of them after convolution. We adopt the hyperbolic tangent function $\phi(x) = 1.7159\tanh(\frac{2}{3}x)$ \citep{IEEELeCun98} as the activation function. The second layer takes the output of the first layer as input, and applies max-pooling over non-overlapping regions of size $2 \times 2$ for each feature map. The third layer adopts filters of size $4 \times 4$, and has $16$ feature maps. The fourth layer then applies max-pooling over non-overlapping pooling regions of size $3 \times 3$. The fifth layer employs filters of size $3 \times 3$ and includes $32$ feature maps. The sixth layer employs $3 \times 3$ non-overlapping max-pooling to the output maps of the fifth layer. After that, the resulting $32$ feature maps of size $3 \times 3$ are cascaded and passed to the first fully-connected layer containing $150$ neurons.

When a cell image is fed into the network, the spatial resolution of each feature map decreases as the features are extracted hierarchically from one layer to next. The spatial information of each cell is extracted by the feature maps because of the spatial convolution and pooling operations, which are important to distinct different staining pattern types. The features obtained are invariant to small translation or shift of cell images, because the filter weights of the convolutional layers are uniform for different regions of the input maps and max-pooling is robust to small variations.

\subsection{Image Preprocessing} \label{subsec:image-pre}
An appropriate image preprocessing method that takes the characteristic of images into consideration is necessary for deep CNNs to obtain good internal feature representation and classification performance.

\begin{figure}[h]
\begin{center}
\includegraphics[width=9cm]{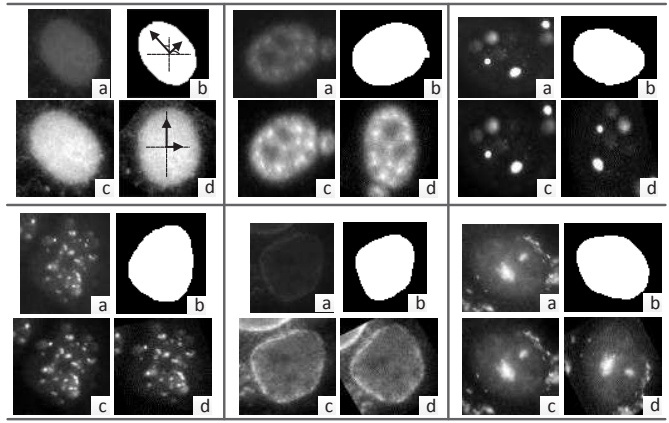}
\caption{Example cells of six classes in ICPR2014 dataset and their corresponding preprocessed and aligned images. There are four images for each cell: (a) the original image; (b) the mask of this cell image (we do not take advantage of it for training the CNN); (c) the preprocessed image when the original image is applied contrast normalization and resized; (d) the aligned image when the contrast normalized image is aligned by PCA and then resized.}
\label{fig:alignedcell}
\end{center}
\end{figure}

The brightness and contrast of the HEp-$2$ cell images provided by the ICPR 2014 contest (ICPR2014 dataset in short) vary greatly. To reduce this variance and enhance the contrast, we normalize each image by first subtracting the minimum intensity value of the image. The resulting intensity is then divided by the difference between the maximum and minimum intensity values. Furthermore, each image is resized to $78 \times 78$ to guarantee a uniform scale of all the images used for training. This size is approximately the average size of all the cell images. Examples of six staining patterns in ICPR2014 dataset and the preprocessed images are shown in Fig. \ref{fig:alignedcell}. In addition, we just use the preprocessed whole cell images to train our network instead of adopting a mask to only keep the foreground within each cell as Malon et al. in \citep{Benchmarking2012}, because the mask information of each cell is usually unavailable in practice, and we find that the classification performance of our system is adversely affected by using cell masks.  

\subsection{Data Augmentation} \label{subsec:data-aug}
Deep CNNs are high-capacity architecture having a large number of parameters to be learned. It will be difficult to effectively train a CNN when training images are insufficient. Data augmentation \citep{ImageNet12} has been regarded as a simple and effective way to generate more samples to train a CNN and gain robustness against a variety of variances.

For data augmentation in the cell image classification, we identify the following two points: i) generating new training images by rotating existing ones can effectively boost the classification performance of the CNNs; ii) instead of merely increasing the robustness of the CNNs against the global orientation of a cell, the extra samples generated via such rotation-based augmentation help to show the intrinsic distribution of the staining patterns belonging to each cell category, which is a more important factor contributing to the improvement of the classification performance.

To demonstrate the first point, we keep rotating each training image with respect to its center by a step of $\theta$ degree. The newly generated images inherit the class label from the original training image, because rotating a cell image does not change its class label. 
By doing so, the original training set is enlarged by a factor of $m = \frac{360}{\theta}$, and this augmented training set is used to train the CNN. 

To demonstrate the second point, we pre-align each cell image to approximately have the same global orientation. In this way, if the global orientation variance is really the main factor affecting the training performance of the CNN, we shall observe some improvement by using the pre-aligned training set. Also, augmenting this pre-aligned training set with rotated images shall not lead to significantly better classification performance. 

To investigate our hypothesis, we apply principal component analysis (PCA) to each cell's mask to obtain the principal direction of its shape. Each contrast normalized cell is rotated to make this principal direction to be vertical and then is resized. Applying this process to all training cell images makes them pre-aligned. These operations are illustrated in the upper left portion (as indicated) in Fig.\ref{fig:alignedcell}, followed by more examples of cell images before and after alignment. After that, we use the pre-aligned training images to train the CNN and then classify test images which are also pre-aligned. 

We find that the CNN trained in this manner does not show better performance than the CNN trained with the preprocessed training images without alignment. However, when data augmentation is applied to the pre-aligned training set images, the performance of the trained CNN increases greatly. This indicates that, in terms of cell classification, adequately demonstrating the staining patterns within a cell image is more important than removing the global orientation variance\footnote{A good example in contrast is human facial image, for which pre-alignment is generally helpful for recognition. This is because the patterns within a facial image, e.g., eyes, nose and mouth, have a rigid geometric association with the global orientation of the face. Pre-aligning the faces with respect to their global orientations effectively makes the patterns inside align with each other. Nevertheless, it is not such a case for cell images.}. Detailed experimental results will be presented in Section \ref{sec:experiment}. 

\subsection{Network Training} \label{subsec:net-training}
Due to the non-convex property of the cost surface of CNNs, it is essential to select appropriate network training parameters, e.g., learning rate, and regularization methods, e.g., weight decay and dropout \citep{hinton2012improving} to make the network converge to good solutions fast.

Our deep CNN is parameterized by the weights and biases of different convolutional layers and fully-connected layers $\{\textbf{W}^{l}, \textbf{b}^{l}\}$, where $l = 1, 3, 5, 7, 8$. The total number of parameters is over $50,000$. The network is trained by minimizing the cross-entropy between the output probability vector ${\boldsymbol{\hat y}} = [\hat y_1, \hat y_2,..., \hat y_n]^\top$ and the binary class label vector ${\boldsymbol{y}} = [y_1, y_2, ..., y_n]^\top$ with one non-zero entry ``$1$'' corresponding to the true class, which is expressed as follows.
\begin{equation}
E({\boldsymbol{y}}, {\boldsymbol{\hat y}}) = - \sum_{j=1}^{n}y_j \log(\hat y_j)
\end{equation}
The weights are initialized from a uniform distribution and the biases are initialized to zero. All these trainable parameters are updated periodically via stochastic gradient descent (SGD) \citep{IEEELeCun98} after evaluating the cost function. Let $w^l$ denote a weight of the $l$th layer, i.e., an element of $\textbf{W}^{l}$. Let $b^l$ be a bias of the $l$th layer (an element of $\textbf{b}^{l}$). Each weight $w^l$ and bias $b^l$ are updated by the following rules:
\begin{equation}
w^l := w^l - \eta \cdot {\frac{{\partial} {E}}{\partial w^l}};\quad b^l := b^l - \eta \cdot {\frac{\partial E}{\partial b^l}}
\end{equation}
where $\eta$ is the learning rate, and $\frac{{\partial} {E}}{\partial w^l}$ and $\frac{{\partial} {E}}{\partial b^l}$ are the partial derivatives of the cost function with respect to $w^l$ and $b^l$ respectively. They are calculated and updated via back-propagating the output error to the $l$th layer \citep{lecun1989backpro} after a number of training images (a mini-batch \citep{Bengio12}) feed into the network.  

To smooth the directions of gradient descent and make the network converge fast, we employ momentum \citep{Bengio12} to speed up the learning by guiding the descent direction with past gradients. The update rules of $w^l$ and $b^l$ become as the follows:
\begin{eqnarray} 
v_w^l &:=& \alpha \cdot v_w^l - \beta \cdot \eta \cdot w^l - \eta \cdot {\frac{{\partial} {E}}{\partial w^l}};\quad w^l := w^l + v_w^l \nonumber   \\
v_b^l &:=& \alpha \cdot v_b^l - \eta \cdot {\frac{{\partial} {E}}{\partial b^l}};\quad b^l := b^l + v_b^l  \nonumber \\
\end{eqnarray} 
where $v_w^l$ and $v_b^l$ are the momentum variables for $w^l$ and $b^l$ respectively; $\alpha$ and $\beta$ are the coefficients of momentum term and weight decay term, and their optimal values are experimentally tuned, as shown in Section \ref{sec:experiment}.  When training error rate becomes stabilized, the learning rate $\eta$ will be reduced to achieve finer learning. The whole training process terminates after the classification error rates of both training set and validation set (which is held out from the given training images) plateau at some epochs. 

In addition, another newly developed regularization strategy, dropout \citep{hinton2012improving}, is also investigated in the network training. It randomly sets a fraction of the activations in the hidden layers to zero to force the hidden units to learn more independent and robust features that could generalize well and to prevent overfitting.

\subsection{Feature Extraction and Classification} \label{subsec:feaextrac-classify}
When classifying a test image, the same preprocessing and rotation in Section \ref{subsec:image-pre} and \ref{subsec:data-aug} are applied. This results in $m$ rotated variants in total. Each of them is forward-propagated through the network, and the probability of this image for each of the $n$ classes is obtained. To further improve the robustness of classification, we select four similar CNNs after the training process becomes stable and use them collectively for classification following \citet{ImageNet12}. The predicted class is the one having the maximum output probability averaged over the $4m$ probabilities, that is,
\begin{equation}\label{eqn:joint-classify}
\hat {l} = \arg \max_j \hat {y}_j =  \arg \max_j \frac{1}{4m} \sum_{k=1}^{m} \sum_{i=1}^{4} \hat {y}_{ik}, ~j = 1,2,..., n.
\end{equation}

\section{Experimental Results} \label{sec:experiment}
We evaluate our CNN classification system on two datasets of HEp-$2$ cell classification competition held by ICPR 2014 and 2012. The evaluation criterion is the mean class accuracy (MCA) newly adopted by ICPR 2014 competition. It is the average of the per-class accuracies \citep{ICPR2014Report} defined as follows: 
\begin{equation} \label{eqn:mca}
\text{MCA} = \frac{1}{n} \sum_{k=1}^{n}\text{CCR}_{k} 
\end{equation} 
where $\text{CCR}_{k}$ is the classification accuracy of class $k$ and $n$ is the number of cell classes.

The average classification accuracy (ACA), which is the overall correct classification rate of all the cell images, used by the previous competition is also calculated for the ease of comparison. 

\subsection{Introduction of the HEp-2 Cell Datasets}
\begin{figure}[H]
\begin{center}
\includegraphics[width=9.5cm]{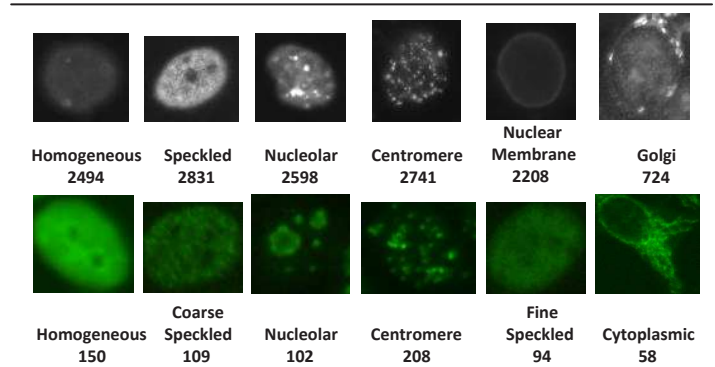}
\caption{Comparison of HEp-$2$ cell images of ICPR2014 dataset (top row) and ICPR2012 dataset (bottom row). The number below the name of each cell is the total number of this kind of cells in the training set of each dataset.}
\label{fig:twodataset}
\end{center}
\end{figure}

\textbf{ICPR2014 cell dataset.}~This dataset contains $13,596$ training cell images, and the test set is reserved by the competition organizers and not published yet. The cell images are extracted from $83$ specimen images captured by monochrome high dynamic range cooled microscopy camera fitted on a microscope with a plane-Apochromat $20\times$/$0.8$ objective lens and an LED illumination source \citep{ICPR2014Report}. These specimen images have been automatically segmented by using the DAPI channel and manually annotated by specialists. Each image belongs to one of the six staining patterns: \textit{Homogeneous, Speckled, Nucleolar, Centromere, Nuclear Membrane} and \textit{Golgi}, as shown in the top row of Fig. \ref{fig:twodataset}.

\textbf{ICPR2012 cell dataset.}~It consists of $1,455$ cell images extracted from $28$ specimens, which are acquired with a fluorescence microscope ($40$-fold magnification) coupled with $50$W mercury vapor lamp and with a digital camera \citep{Benchmarking2012}. The dataset is pre-partitioned into training set ($721$ images) and test set ($734$ images). Each image belongs to one of the six classes: \textit{Homogeneous, Coarse Speckled, Nucleolar, Centromere, Fine Speckled} and \textit{Cytoplasmic}, as shown in the bottom row of Fig. \ref{fig:twodataset}. 

Comparing the two datasets shows that two of the six classes are different. Specifically, two sub-categories of ICPR2012 dataset (\textit{Fine Speckled} and \textit{Coarse Speckled}) are merged into one category (\textit{Speckled}) in ICPR2014 dataset, and two less frequent staining patterns appearing in daily clinical cases, \textit{Golgi} and \textit{Nuclear Membrane} are introduced in ICPR2014 dataset for developing more realistic HEp-$2$ cell classification systems. \textbf{Moreover, because the images in the two datasets are captured with different laboratory settings, a classification system that can be easily transferred from one dataset to the other one will be highly desired.}
\subsection{Experiments of Hyper-parameters Optimization}
This experiment demonstrates the importance of properly tuning the hyper-parameters in the CNN-based system. We categorize the hyper-parameters into two groups: model-relevant and training-relevant, as listed in Tables \ref{Table:model-para} and \ref{Table:train-para}.
\begin{table}[H]
\centering
\resizebox{\linewidth}{!}{\begin{tabular}{c c l}
\hline
Layer Number & Layer Type & Hyper-parameter \\
\hline
Layer 1 & Convolution  & Filter size: $7 \times 7$\\
        &              & Feature map number: $6$ \\  
        &              & Activation function: \\
        &              & hyperbolic tangent $\phi(x) = 1.7159\tanh(\frac{2}{3}x)$\\
\hline
Layer 2 & Pooling & Pooling region size: $2 \times 2$ \\ 
        &         & Pooling method: max-pooling \\
\hline
Layer 3 & Convolution  & Filter size: $4 \times 4$\\  
        &              & Feature map number: $16$ \\  
        &              & Activation function: \\
        &              & hyperbolic tangent $\phi(x) = 1.7159\tanh(\frac{2}{3}x)$\\
\hline
Layer 4 & Pooling & Pooling region size: $3 \times 3$ \\ 
        &         & Pooling method: max-pooling\\
\hline
Layer 5 & Convolution  & Filter size: $3 \times 3$\\  
        &              & Feature map number: $32$ \\  
        &              & Activation function: \\
        &              & hyperbolic tangent $\phi(x) = 1.7159\tanh(\frac{2}{3}x)$\\
\hline
Layer 6 & Pooling & Pooling region size: $3 \times 3$ \\
        &         & Pooling method: max-pooling\\
\hline
Layer 7 & Full connection & Neurons number: $150$ \\ 
        &                 & Activation function: \\
        &                 & hyperbolic tangent $\phi(x) = 1.7159\tanh(\frac{2}{3}x)$\\
\hline
\end{tabular}}
\caption{Model-relevant hyper-parameters obtained}\label{Table:model-para}
\end{table}

\begin{table}[H]
\centering
\resizebox{\linewidth}{!}{\begin{tabular}{c c c c c c}
\hline
Hyper-parameter &  \tabincell{c}{Initial \\learning rate} & \tabincell{c}{Mini-batch\\ size}& \tabincell{c}{Momentum\\coefficient} & \tabincell{c}{Weight decay\\coefficient} & Dropout ratio \\
\hline
Value & 0.01 & 113 & 0.9 & 0.0005 & 0 \\
\hline
\end{tabular}}
\caption{Training-relevant hyper-parameters obtained}\label{Table:train-para}
\end{table}

To tune these hyper-parameters, we randomly partition the $13,596$ cell images of ICPR2014 dataset into three subsets, that is, $64$\% for training ($8701$ images), $16$\% for validation ($2175$ images), and  $20$\% for test ($2720$ images). This partition is utilized by all experiments on ICPR2014 dataset (multiple partitions could be certainly implemented when the computational resource is not an issue.). Data augmentation is not used when tuning hyper-parameters. Following \citet{Bengio12}, the parameters are tuned until the error rate of not only the training set but also the validation set become sufficiently small and stabilized. The hyper-parameters obtained by this tuning process are summarized in Tables \ref{Table:model-para} and \ref{Table:train-para}. 

\begin{figure}[H]
\centering
\subfloat[Learning rate = 0.001]{\includegraphics[width=4.5cm]{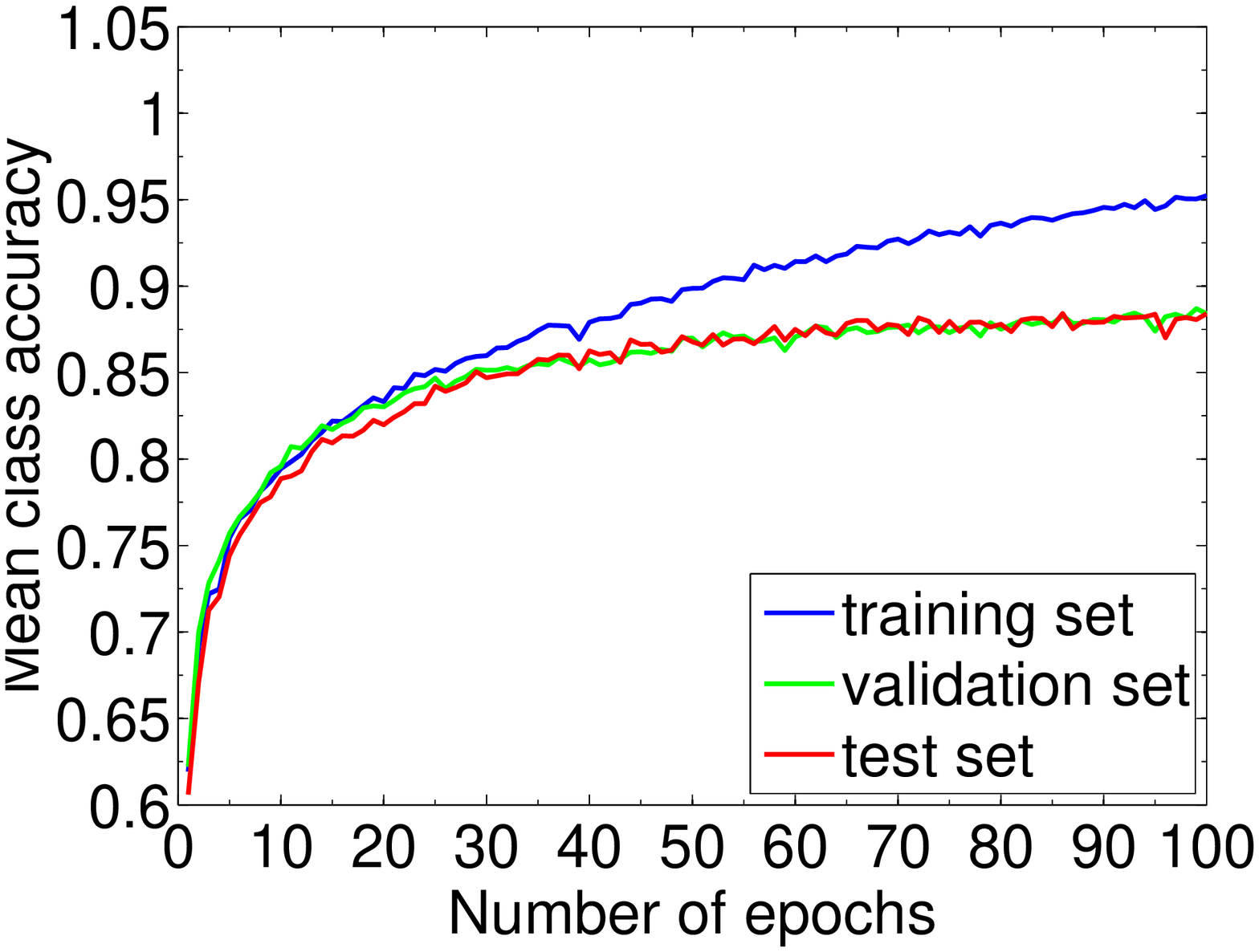}%
\label{subfig:lr1}}
\hfil
\subfloat[Learning rate = 0.01]{\includegraphics[width=4.5cm]{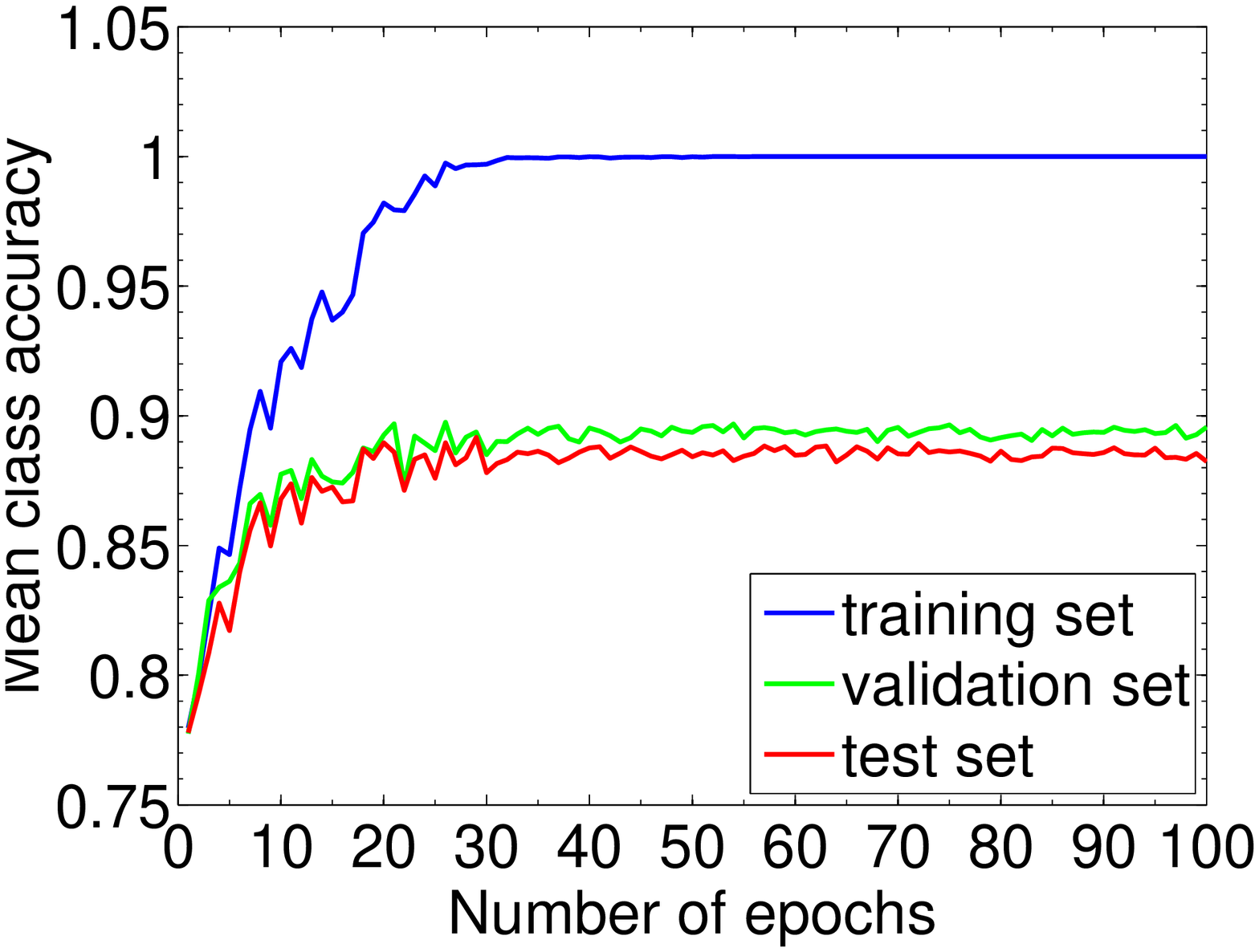}%
\label{subfig:lr2}}
\hfil
\subfloat[Learning rate = 0.1]{\includegraphics[width=4.5cm]{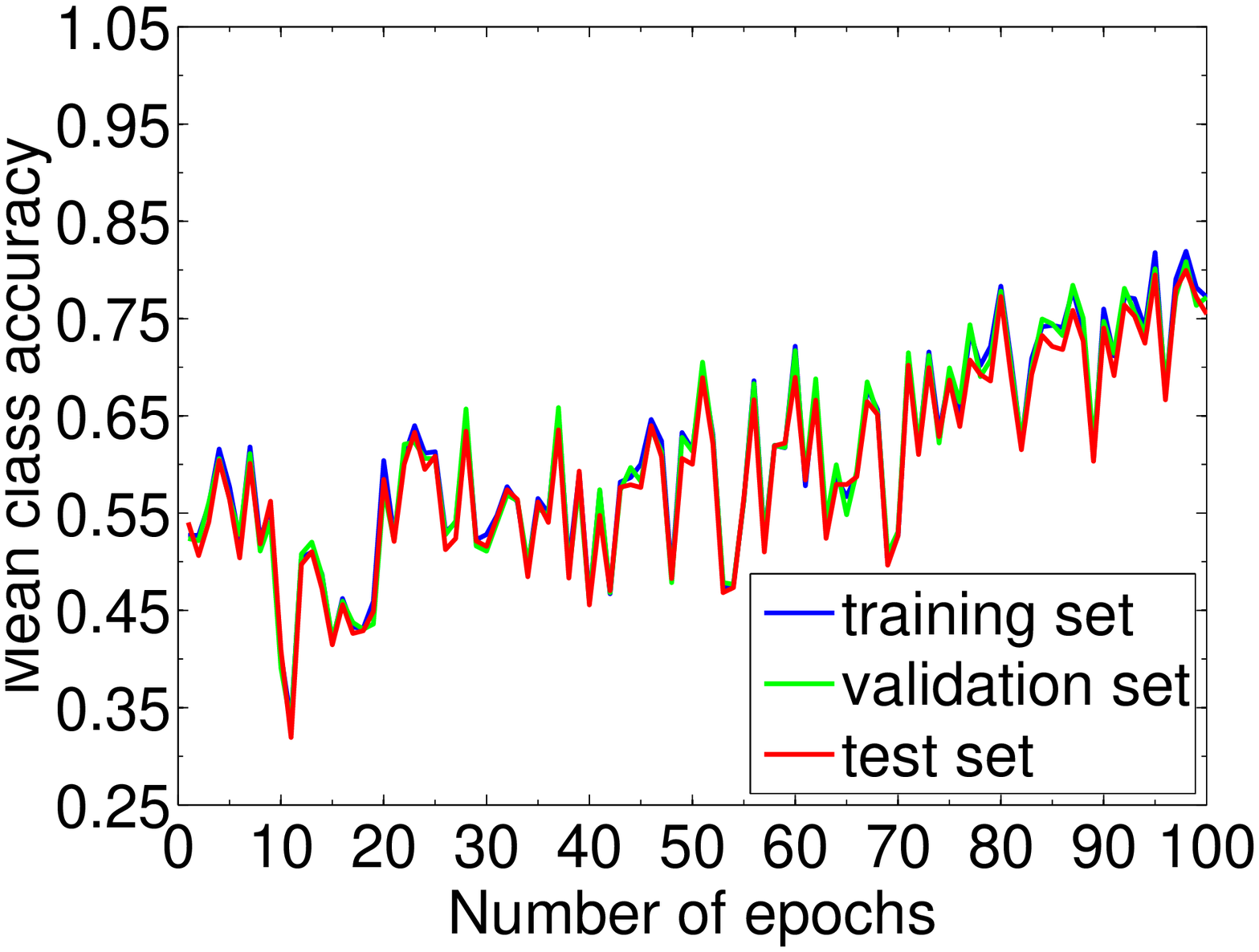}%
\label{subfig:lr3}}
\caption{Demonstration of the impact of learning rate. It shows that an over-small learning rate, e.g., $0.001$, slows down the learning process, whereas an over-large learning rate, e.g., $0.1$, destabilizes the learning process and degrades the classification performance. A better classification result can be obtained by properly tuning the learning rate, as shown in (b).}
\label{fig:learningrate}
\end{figure}

\begin{figure}[H]
\centering
\subfloat[Mini-batch size = 11]{\includegraphics[width=5cm]{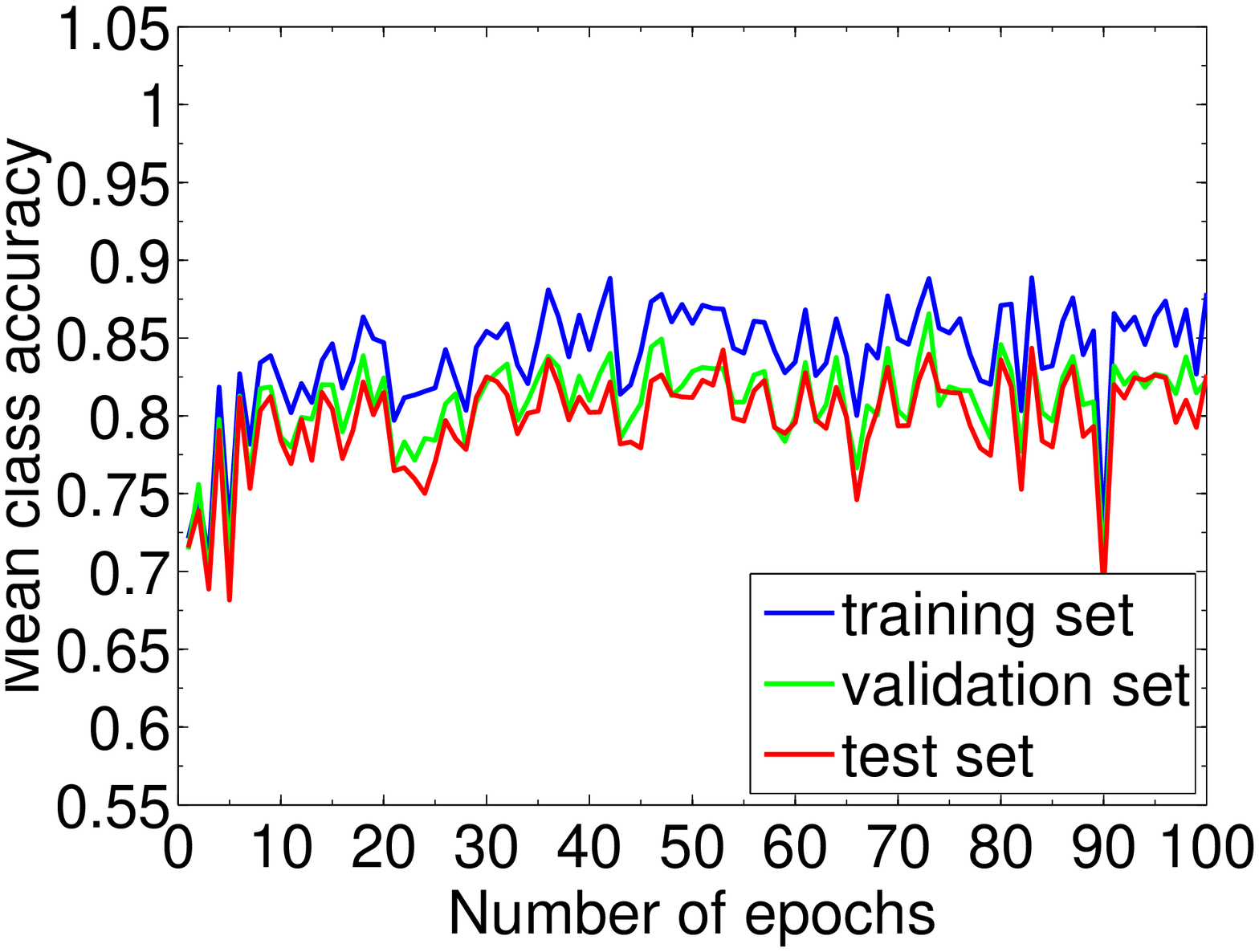}%
\label{subfig:m1}}
\hfil
\subfloat[Mini-batch size = 77]{\includegraphics[width=5cm]{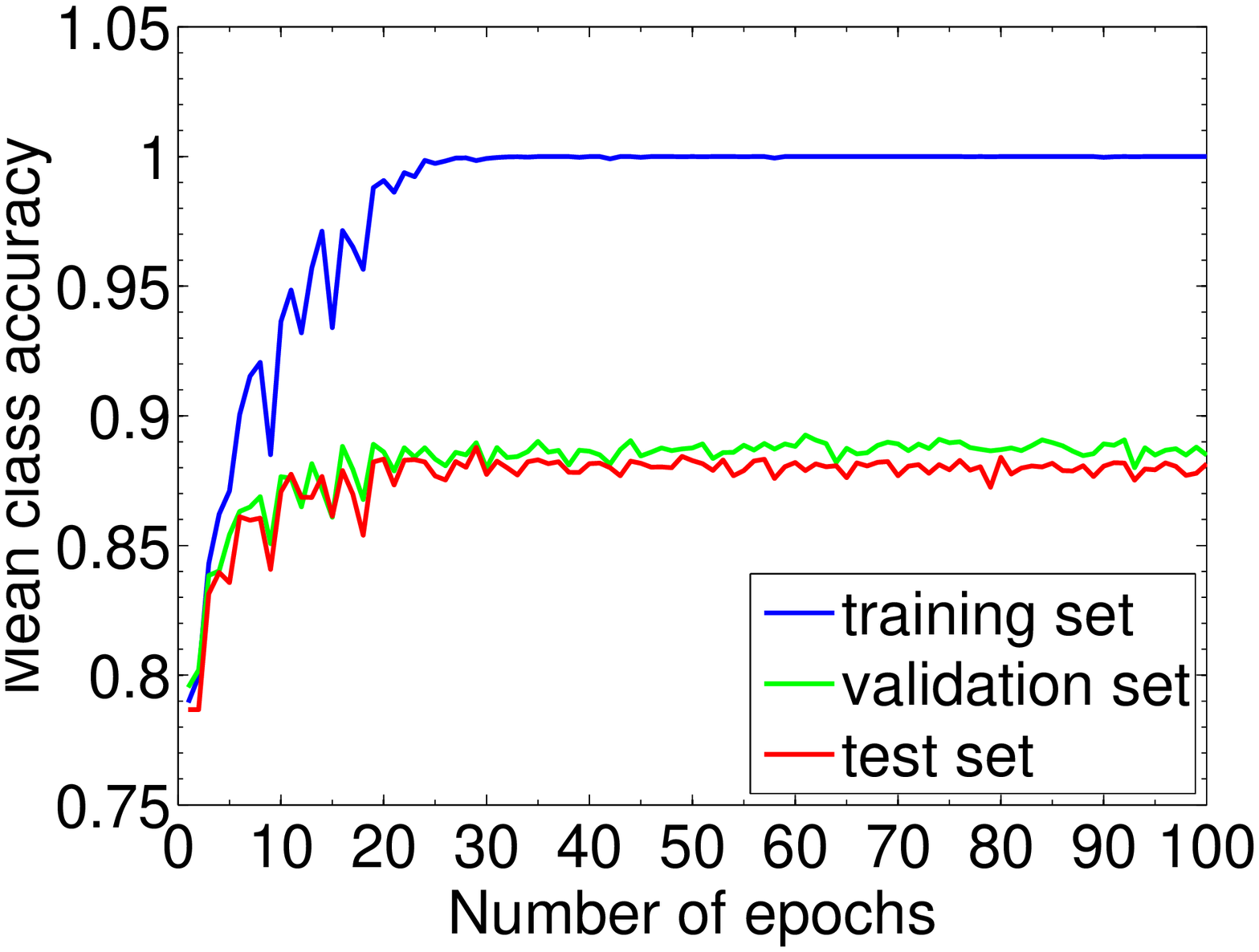}%
\label{subfig:m2}}
\hfil
\subfloat[Mini-batch size = 113]{\includegraphics[width=5cm]{./figure_new/MCA_8701_part0_LR01_new_MIA}%
\label{subfig:m3}}
\hfil
\subfloat[Mini-batch size = 791]{\includegraphics[width=5cm]{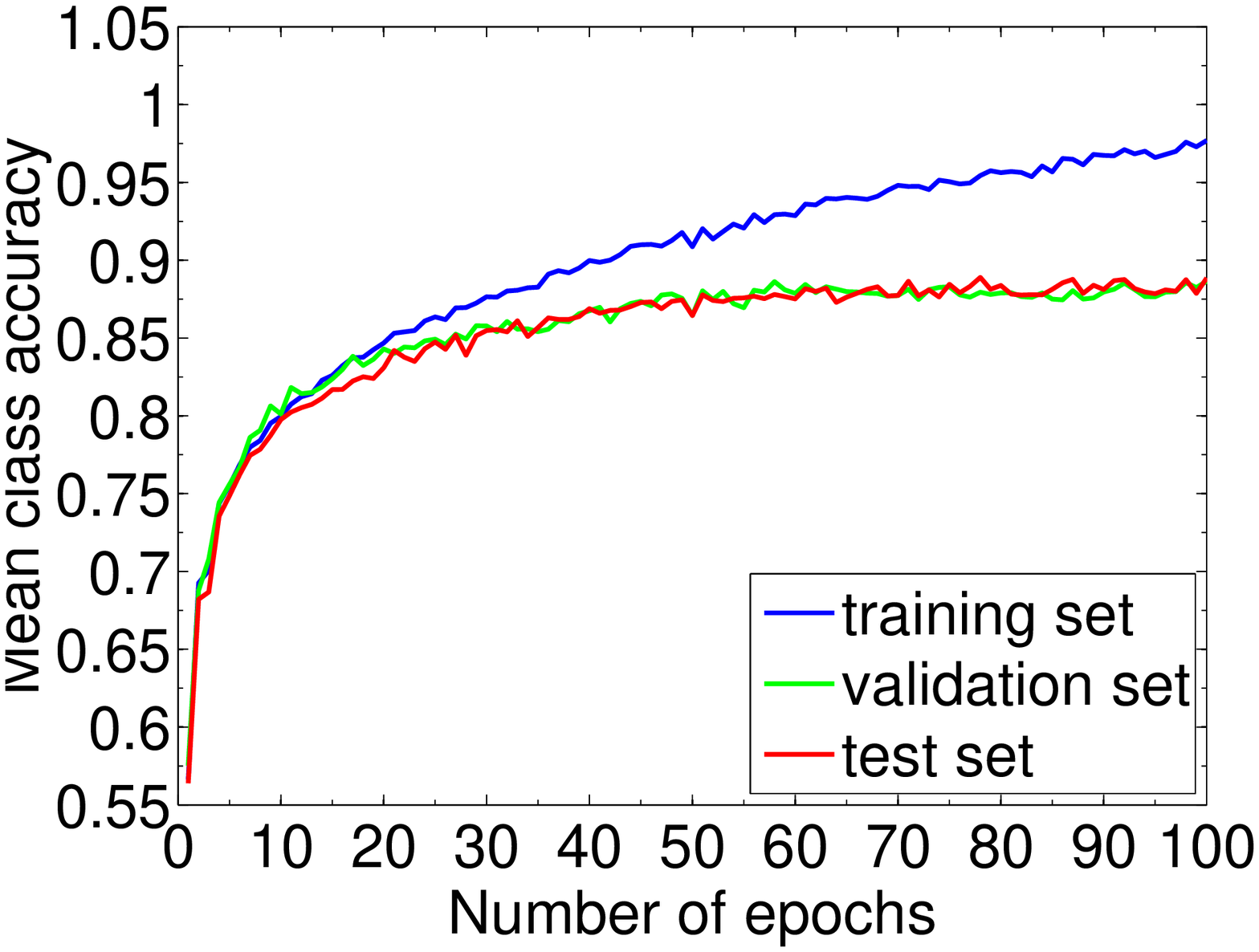}%
\label{subfig:m4}}
\caption{Demonstration of the impact of mini-batch size. It shows that when mini-batch size is unnecessarily small, the learning process becomes bumpy and does not lead to the best result. On the other hand, when the mini-batch size is too large, the learning process becomes less responsive and the learning efficiency is decreased.}
\label{fig:minibatch}
\end{figure}

\begin{figure}[H]
\centering
\subfloat[Momentum coefficient = 0]{\includegraphics[width=5cm]{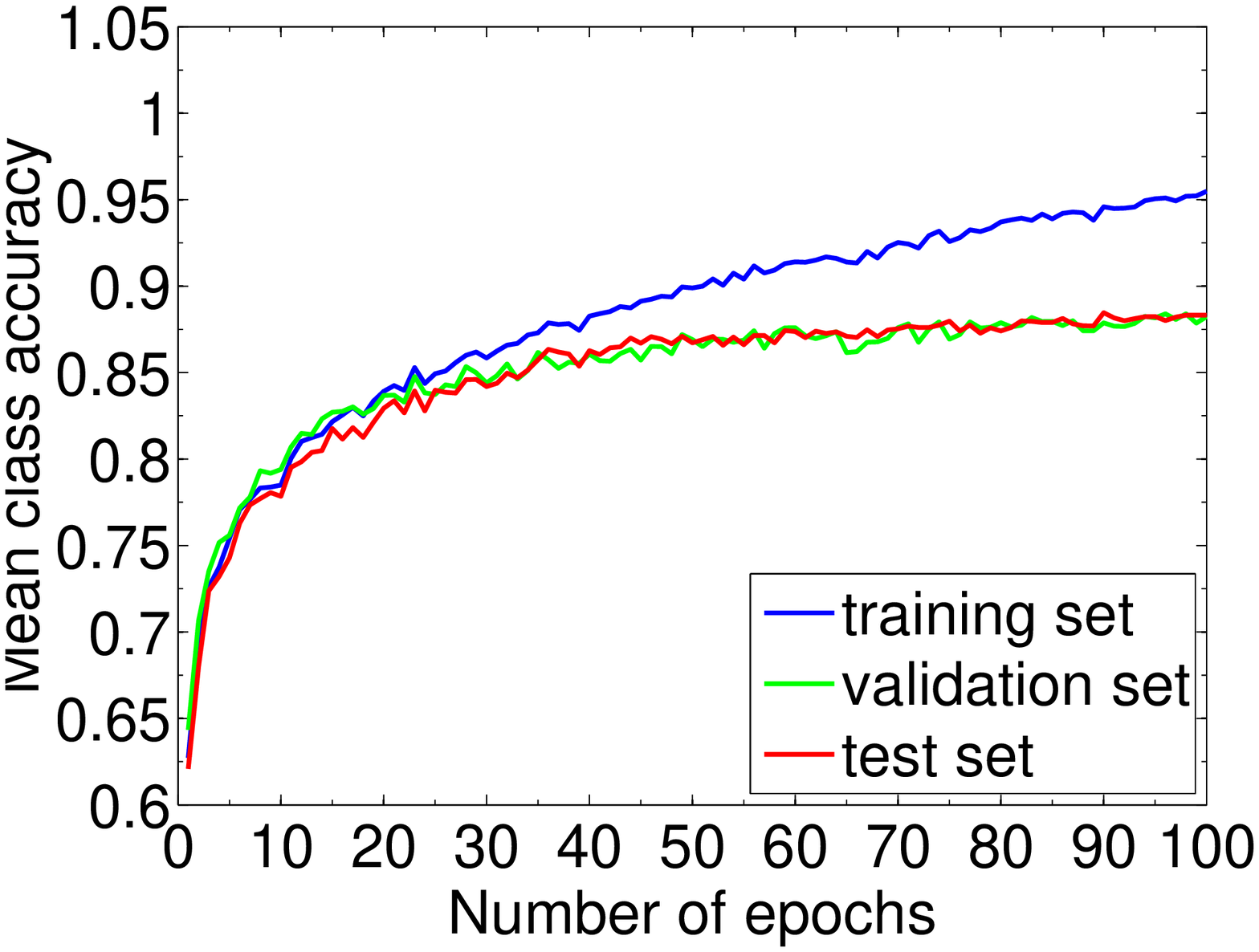}%
\label{subfig:mom1}}
\hfil
\subfloat[Momentum coefficient = 0.8]{\includegraphics[width=5cm]{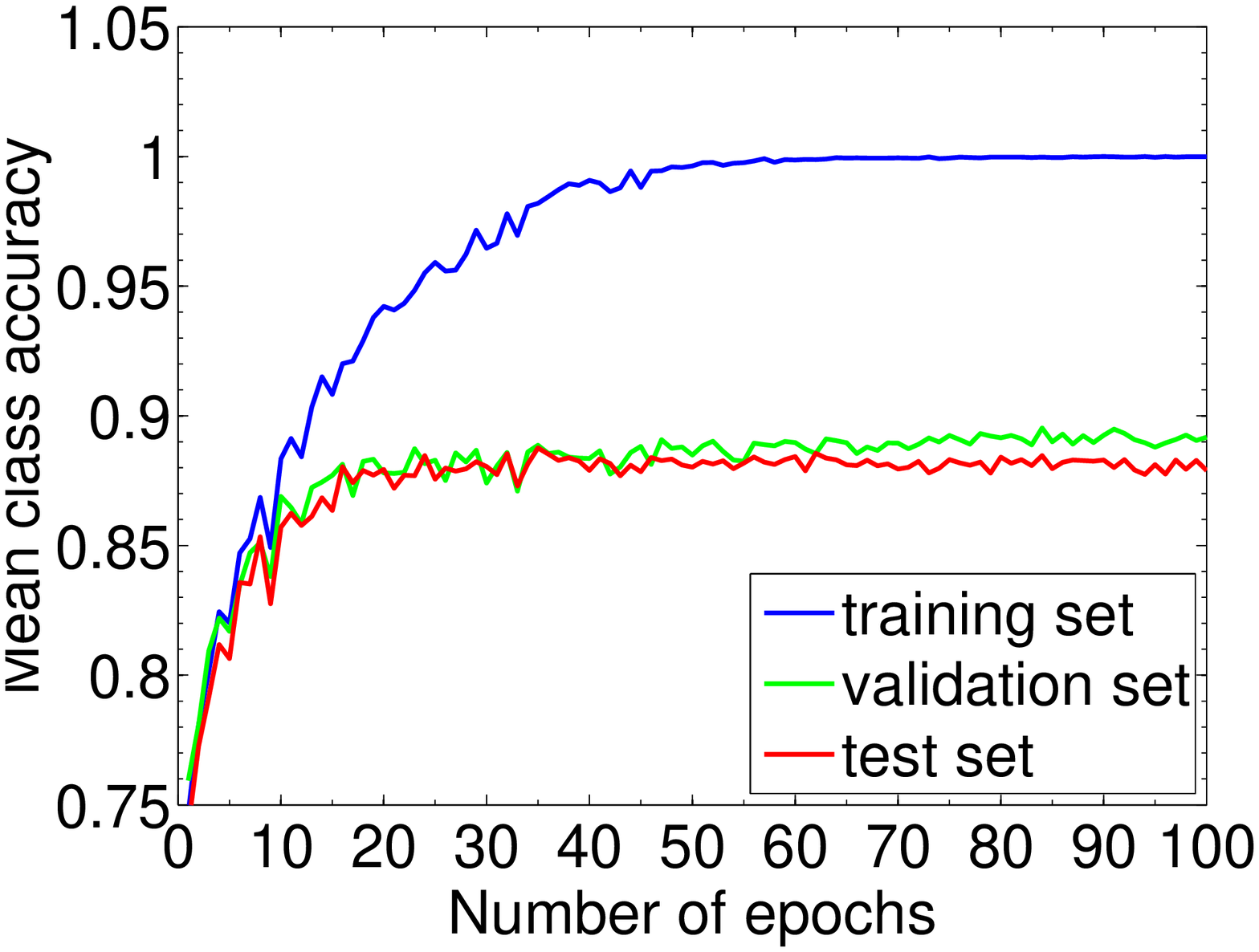}%
\label{subfig:mom2}}
\hfil
\subfloat[Momentum coefficient = 0.9]{\includegraphics[width=5cm]{./figure_new/MCA_8701_part0_LR01_new_MIA}%
\label{subfig:mom3}}
\hfil
\subfloat[Momentum coefficient = 0.97]{\includegraphics[width=5cm]{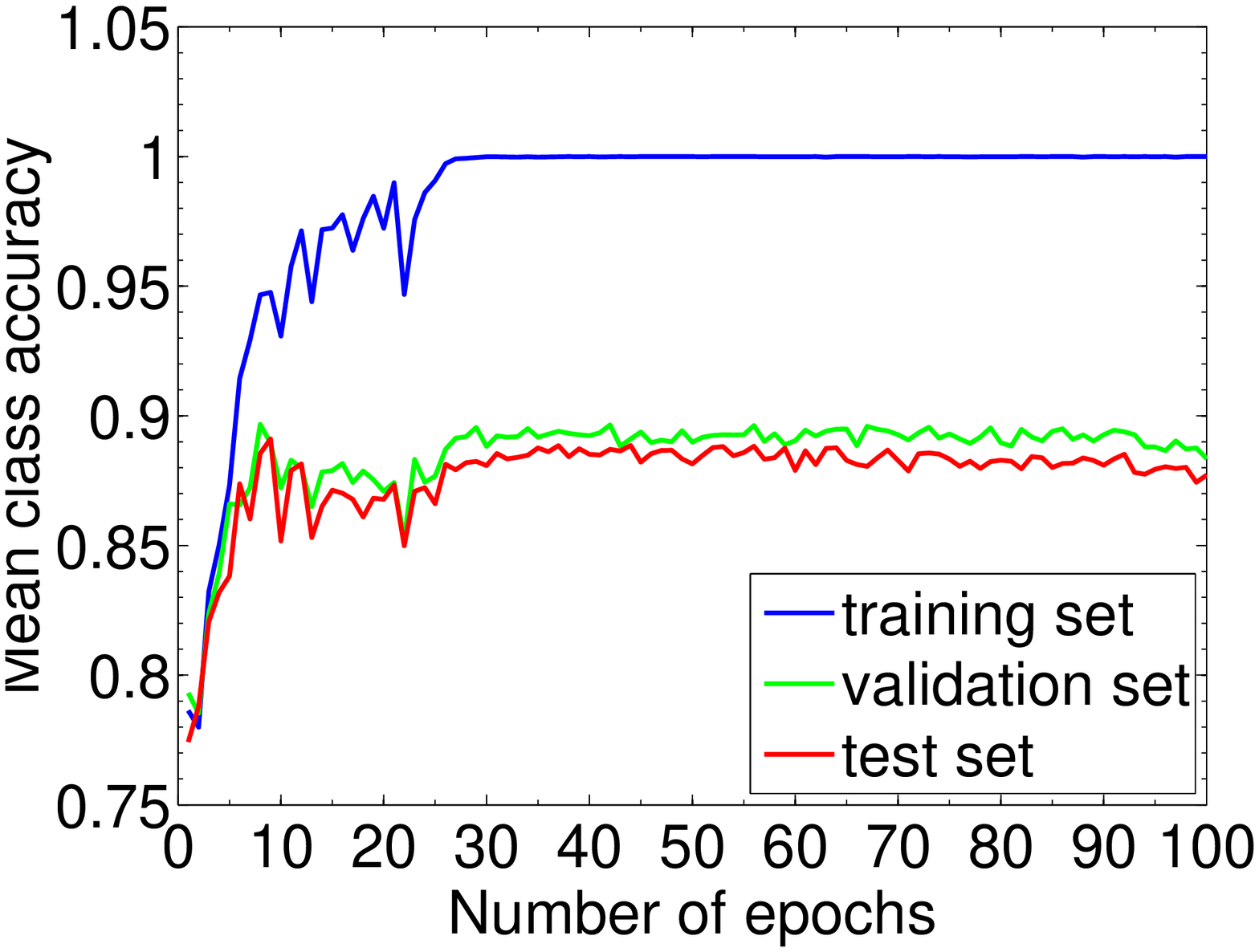}%
\label{subfig:mom4}}
\caption{Demonstration of the impact of momentum. It shows that using momentum can well accelerate the learning process. Meanwhile, a large momentum coefficient, e.g., $0.97$, makes the descent direction dominated by the previous ones and causes oscillation at the initial stage. Also, it decreases the classification performance at the later stage.}
\label{fig:momen}
\end{figure}

\begin{figure}[H]
\centering
\subfloat[Weight decay coefficient = 0.00005]{\includegraphics[width=4.5cm]{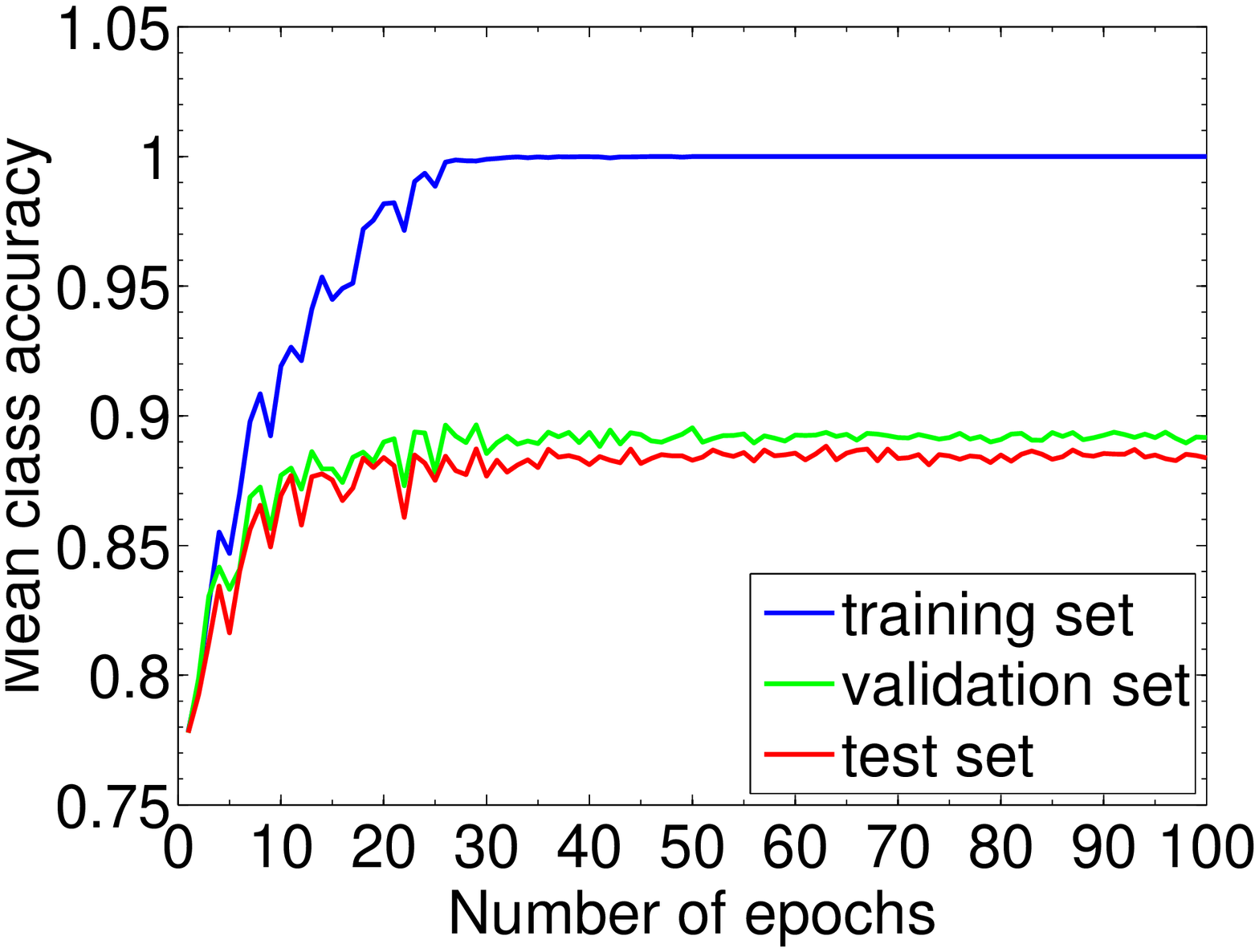}%
\label{subfig:wei1}}
\hfil
\subfloat[Weight decay coefficient = 0.0005]{\includegraphics[width=4.5cm]{./figure_new/MCA_8701_part0_LR01_new_MIA}%
\label{subfig:wei2}}
\hfil
\subfloat[Weight decay coefficient = 0.005]{\includegraphics[width=4.5cm]{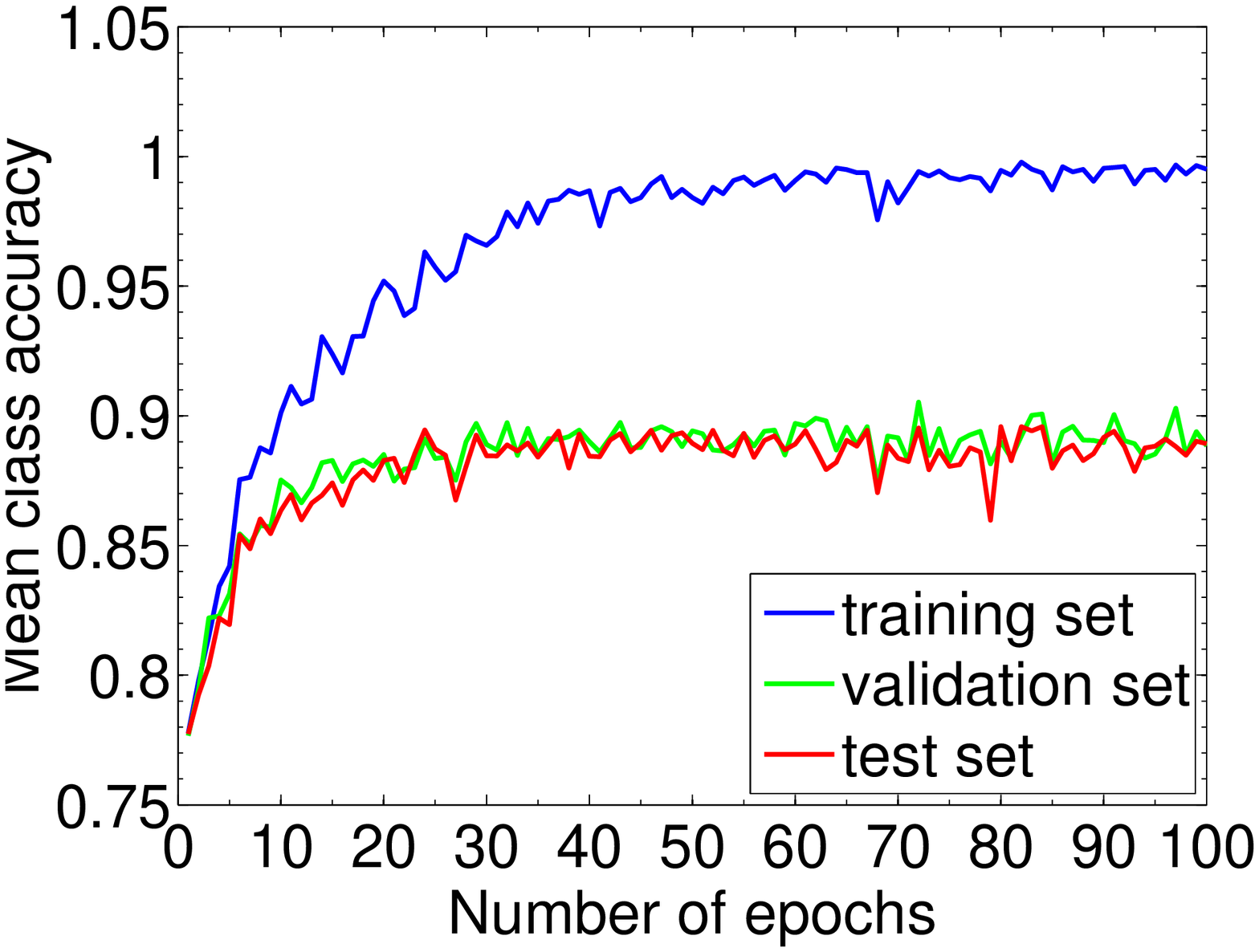}%
\label{subfig:wei3}}
\caption{Demonstration of the impact of weight decay. It shows that a smaller weight decay coefficient seems to be a safer choice, while a larger coefficient, e.g., $0.005$, could destabilize the learning process.}
\label{fig:weight}
\end{figure}

\begin{figure}[H]
\centering
\subfloat[Dropout ratio = 0.5]{\includegraphics[width=5.5cm]{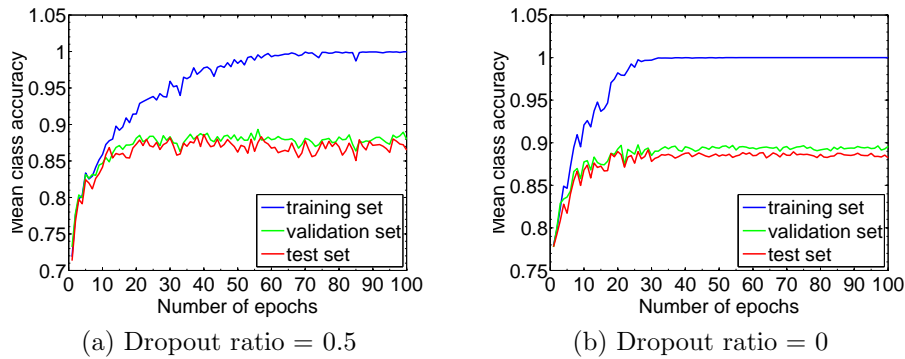}%
\label{subfig:drop1}}
\hfil
\subfloat[Dropout ratio = 0]{\includegraphics[width=5.5cm]{./figure_new/MCA_8701_part0_LR01_new_MIA}%
\label{subfig:drop2}}
\caption{Demonstration of the impact of dropout. It shows that the dropout strategy shall be used cautiously. As seen in (a), the learning process becomes slow and fluctuated on ICPR2014 cell dataset, when dropout is applied. A better learning process is obtained in (b) after removing dropout.}
\label{fig:dropout}
\end{figure}

We highlight that training-relevant hyper-parameters can significantly affect the convergence of cost function, the learning speed and the generalization capability of the network. Their impacts are demonstrated via the learning curves of MCA on training, validation and test sets shown from Fig. \ref{fig:learningrate} to Fig. \ref{fig:dropout}. In each figure, we focus on one hyper-parameter while the others are set to their optimal values in Table \ref{Table:train-para}. 

Fig. \ref{fig:learningrate} \subref{subfig:lr1} indicates that when learning rate is small, e.g., $0.001$, the learning process is so slow that the MCA of the three sets have not become stable in $100$ epochs. Properly increasing the learning rate effectively improves learning efficiency and the MCA becomes stable in $35$ epochs, as shown in Fig. \ref{fig:learningrate} \subref{subfig:lr2}. At the same time, an over-large learning rate, e.g., $0.1$, will destabilize the learning process and degrade the classification performance. Also, Fig. \ref{fig:minibatch}, \ref{fig:momen}  and \ref{fig:weight} demonstrate the impacts of mini-batch size, momentum and weight decay, respectively.

The comparison in Fig. \ref{fig:dropout} shows that the dropout strategy \citep{hinton2012improving} shall be used cautiously. When dropout with ratio of $0.5$ (randomly setting the activations to zero with probability of $0.5$) is applied to the first fully-connected layer of our CNN system, the learning process becomes slow and fluctuated on ICPR2014 cell dataset. A stabler and faster learning process without overfitting on the test set is gained when removing dropout, as well as better classification performance. This indicates that the neurons at the first fully-connected layer may have to work together to distinguish different staining patterns. In light of this, we decide not to employ dropout when training our network on ICPR2014 dataset. 

In sum, among the hyper-parameters of a CNN, the learning rate, mini-batch size, momentum coefficient, and weight decay coefficient can significantly impact the network training process. They have to be carefully tuned before satisfactory classification performance is obtained. For our deep CNN system, with the hyper-parameters set in Table \ref{Table:train-para}, we can achieve the MCA of $89.17$\% on the test set of ICPR2014 dataset without using data augmentation.

\subsection{Experiments on Data Augmentation}
This experiment demonstrates the two points presented in Section~\ref{subsec:data-aug}, which are recapped as follows: i) the performance of the CNN can be greatly boosted by generating new training images via rotation; ii) the extra samples generated via such rotation-based augmentation help to enrich our observations of the staining patterns of each cell category for training the CNN, which is a more important factor contributing to the improvement of the classification performance than increasing robustness of the CNN against the global orientation of cells. 

\textbf{Effectiveness of data augmentation.} We augment the training set by rotating each cell image for $360^\circ$, with the step of $36^\circ$, $18^\circ$ and $9^\circ$, respectively. In this way, the training set is expanded by $10$, $20$ and $40$ times, and they are used to train the CNNs, respectively. To improve the robustness of our system, we select four CNNs corresponding to the $75$th, $85$th, $95$th and $100$th epochs after the network learning becomes stable\footnote{This strategy is adopted as a model average. Different number of CNNs may be chosen, e.g. 3 or 5, to compromise between the computational expense and performance, which leads to similar classification accuracy in our experiments.} as in \citet{ImageNet12}. A test image will go through the same rotation process as the training images and be jointly classified by the four CNNs as in Eq.(\ref{eqn:joint-classify}). This system is named as ``CNN''.  As shown in the first row of Table~\ref{Table:data-aug}, the MCA is significantly improved (by more than $7$ percentage points) from ``No data augmentation'' to ``Augmentation by a rotation angle step of $36^\circ$''. Furthermore, applying a smaller angle step to generate more training data pushes the MCA even higher, reaching $96.76$\%. Similar results can be observed on the ACA values. These consistent and continuous improvements well demonstrate the effectiveness and efficiency of data augmentation on cell image classification. 
\begin{table}[ht]
\centering
\resizebox{\linewidth}{!}{\begin{tabular}{c c c c c c}
\hline
\textbf{Method}  & \tabincell{c}{\textbf{Accuracy}\\(on test set)} & \tabincell{c}{No\\data augmentation} & \tabincell{c}{Augmentation by\\a rotation angle step of $36^\circ$} & \tabincell{c}{Augmentation by\\a rotation angle step of $18^\circ$} & \tabincell{c}{Augmentation by\\ a rotation angle step of $9^\circ$} \\
\hline
\multirow{2}{*}{CNN} & MCA(\%)  & 88.58 & 95.99 &96.71 & \textbf{96.76}  \\ 
                                     & ACA(\%)  & 89.04 & 96.51 &97.10 & \textbf{97.24}  \\
\hline
\multirow{2}{*}{CNN-Align} & MCA(\%) & 88.86 & 95.13 & 96.50 & \textbf{96.52} \\ 
                                              & ACA(\%)  & 88.71 & 95.33 & 96.84 & \textbf{96.84}\\
\hline
\end{tabular}}
\caption{Classification accuracy of our deep CNN on ICPR2014 dataset}\label{Table:data-aug}
\end{table}

\textbf{Data augmentation vs pre-alignment.} To gain more insight on the rotation-based data augmentation, we pre-align all the cell images with PCA as described in Section \ref{subsec:data-aug} to train the CNNs. We call this method ``CNN-Align''. Two experiments are conducted: i) only using these aligned images to train the CNNs without performing data augmentation; and ii) as a comparison, we further rotate each aligned training image by $360^\circ$, also with an angle step of $36^\circ$, $18^\circ$ and $9^\circ$, respectively. The augmented training set is used for training. As previous, augmentation (or no augmentation) is equally applied to test images. 

As shown in Table \ref{Table:data-aug}, when no augmentation is performed, CNN-Align does not achieve any improvement over CNN. This indicates that pre-alignment does not help here. In contrast, when training data are augmented by rotation (even with the largest angle step of $36^\circ$), CNN-Align improves significantly. This sharp change clearly demonstrates that through the rotation-based augmentation, the network can access more examples showing the diverse staining patterns within cell images. This is a more important factor contributing to the performance improvement compared with pre-alignment that only tackles the global orientation variance of cells. 

\begin{figure}[H]
\centering
\subfloat[1st convolutional layer features]{\includegraphics[width=5.5cm]{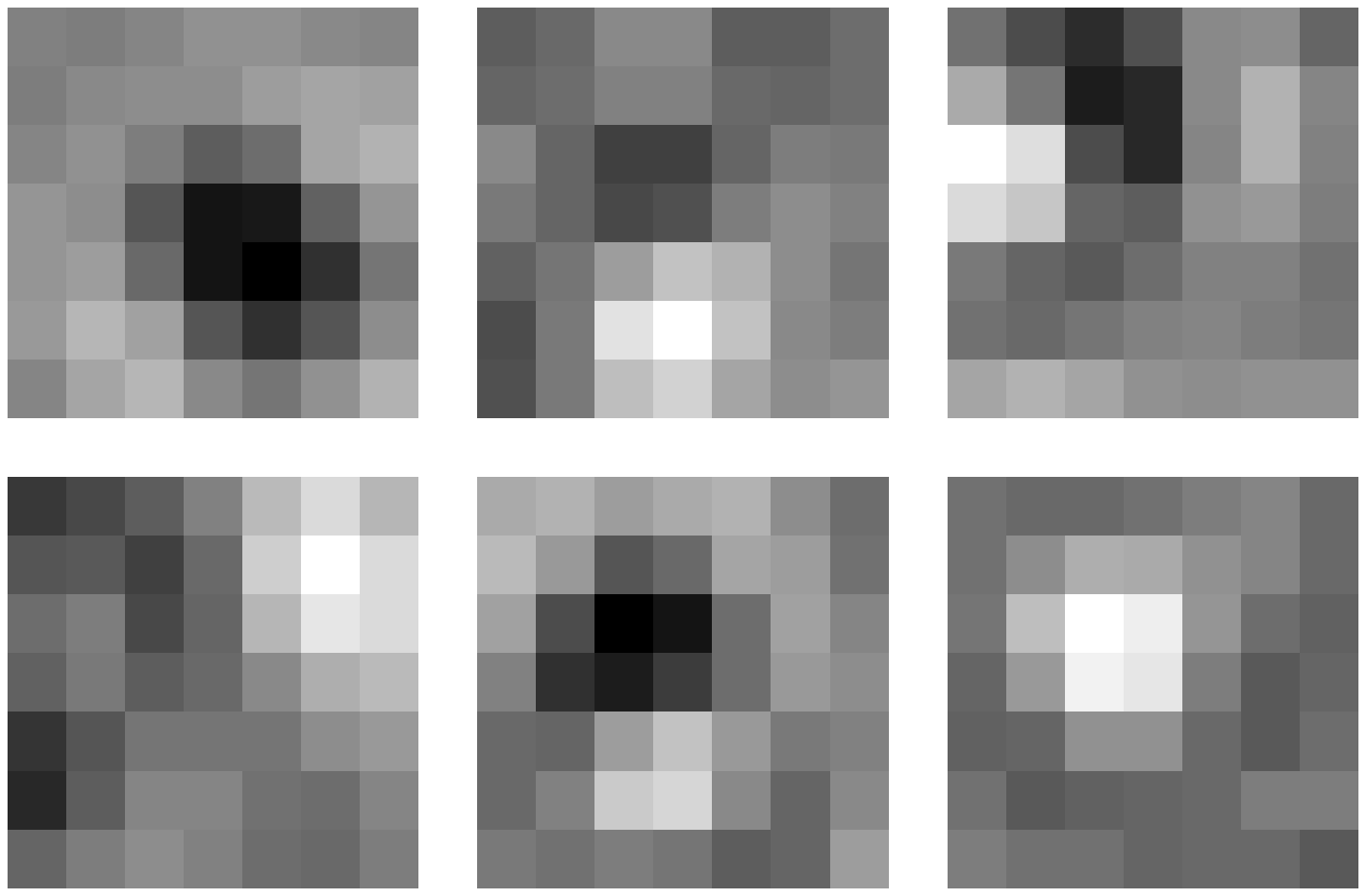}
\label{subfig:drop1}}
\hfil
\subfloat[2nd convolutional layer features]{\includegraphics[width=5.5cm]{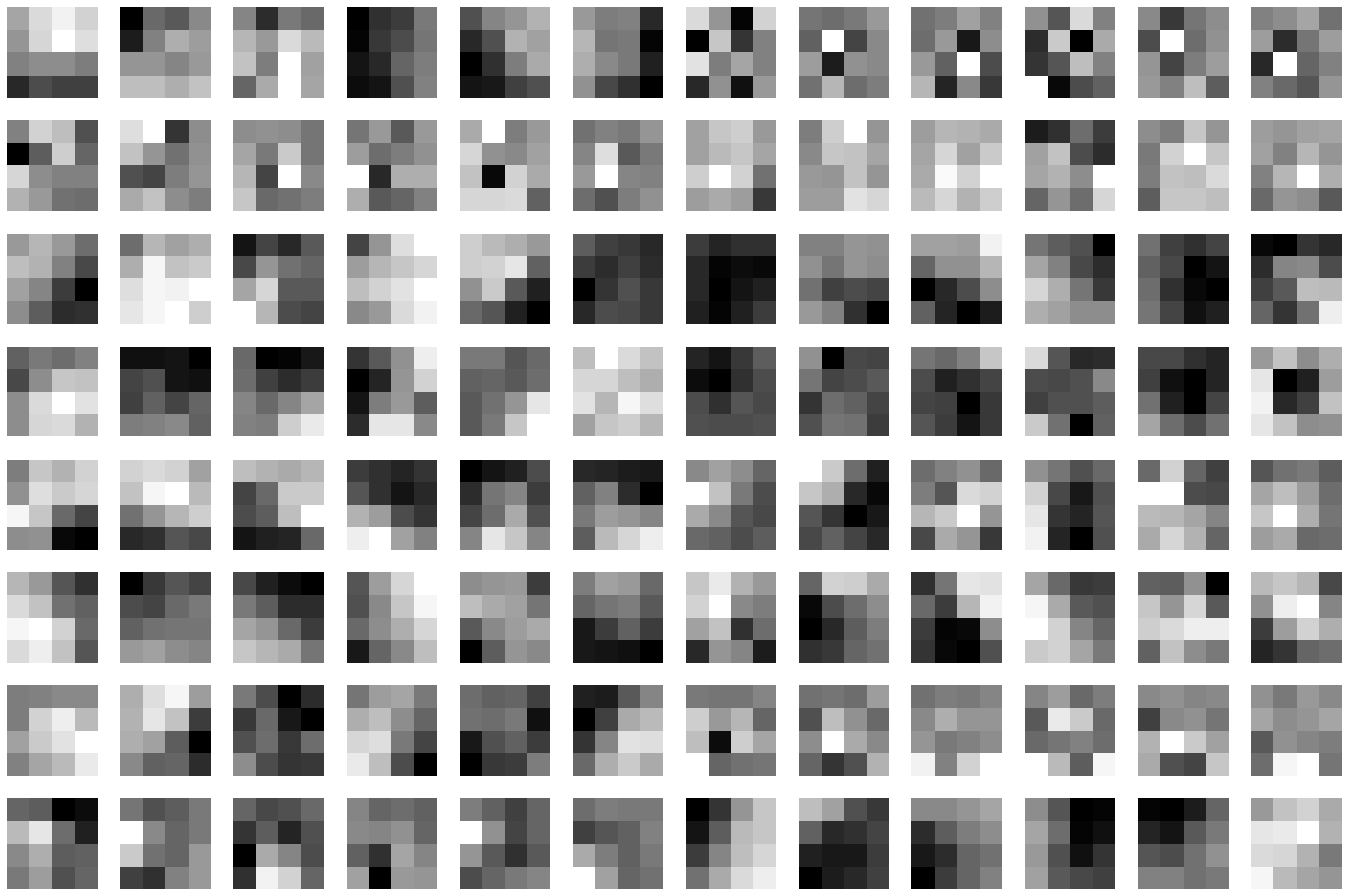}
\label{subfig:drop2}}
\caption{The features learned by the first and second convolutional layers. In general, most of the filters are stain-like texture detectors, and some are edge-like extractors.}
\label{fig:filtervisualize}
\end{figure}

The features (filters) learned by the first and second convolutional layers of CNN corresponding to the $100$th epoch trained with $9^\circ$ rotated cell images are depicted as Fig. \ref{fig:filtervisualize}. It can be seen that the filters of the first convolutional layer are stain-like texture detectors. Some of the second convolutional layer filters are edge-like detectors, and most of them are also stain-like texture extractors. 

\begin{figure}[H]
\begin{center}
\includegraphics[width=9cm]{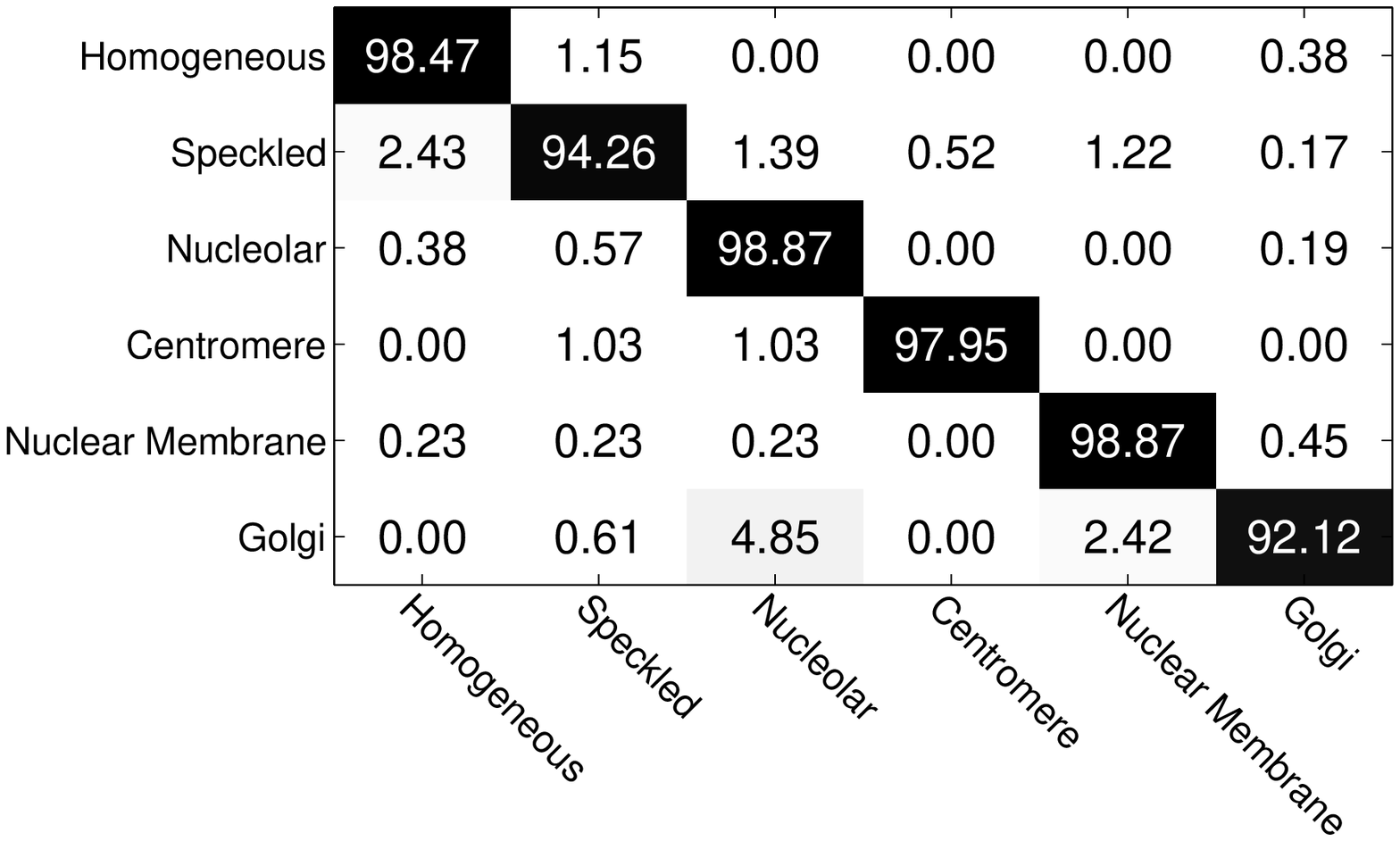}
\caption{Confusion matrix of our best CNN ($9^\circ$ rotation) (\%).}
\label{fig:confusionmatrix}
\end{center}
\end{figure}

\begin{figure}[h]
\begin{center}
\includegraphics[width=7cm]{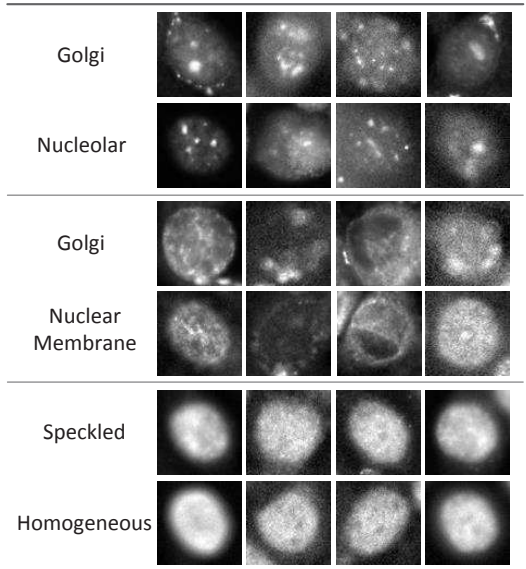}
\caption{Misclassification examples of the three highest misclassification rates in the confusion matrix of Fig. \ref{fig:confusionmatrix}. Every two rows form a group, and the first row shows cells that are misclassified to the cell type of the second row.}
\label{fig:misclassification}
\end{center}
\end{figure}

In addition, the confusion matrix of the best CNN (trained with the rotation angle step of $9^\circ$) is shown in Fig. \ref{fig:confusionmatrix}. The overall classification performance is very promising. The staining patterns \textit{Nucleolar} and \textit{Nuclear Membrane} obtain the highest classification accuracy (both $98.87$\%), which means that they are well separated from the others. The maximum misclassification rate ($4.85$\%) happens to \textit{Golgi} cells. They are easy to be misclassified as \textit{Nucleolar} cells, because both patterns consist of a few large dots within the cells (see misclassification examples in Fig. \ref{fig:misclassification}). Also, \textit{Golgi} can be confused with \textit{Nuclear Membrane}. This may be because when the large dots within \textit{Golgi} cells are at the edge, they will look like the \textit{Nuclear Membrane} cells having ring-like edges. In addition, the \textit{Speckled} cells are easy to be misclassified as \textit{Homogeneous} cells, probably because the densely distributed speckles are the main signatures for both patterns. Misclassification examples of these staining patterns are shown in Fig. \ref{fig:misclassification}.

\subsection{Comparison with the BoF and Fisher Vector Models}
\textbf{Experimental setting.} To ensure a fair comparison, the same image preprocessing in our CNN model is equally used in both models. For each cell image, SIFT descriptors are extracted from densely sampled patches with a stride of two pixels. The visual dictionary is generated by applying the $k$-means clustering to the descriptors extracted from training images. Local soft-assignment coding (LSC) \citep{van2008kernel, liu2011defense} is employed to encode the SIFT descriptors. SPM is used to partition each image into $1 \times 1$, $2 \times 2$ and $1 \times 3$ regions, and max-pooling is applied to extract representations from each region.  

A similar setting is applied to the FV model. In addition, the $128$-dimensional SIFT descriptors are decorrelated and reduced to dimensions of $64$ by PCA as in \citet{sanchez2013image}. A GMM is then estimated to represent the visual dictionary. Afterwards, each PCA-reduced SIFT descriptor is encoded with the improved Fisher encoding \citep{perronnin2010improving}, where the signed square-root and $l^2$-normalization are applied to the coding vector. SPM with four regions ($1 \times 1$ and $1 \times 3$) are adopted \citep{sanchez2013image}. Following the literature, a multi-class linear SVM classifier is used in the BoF and FV models. In our implementation of BoF and FV, the publicly available VLFeat toolbox \citep{vedaldi2010vlfeat} is used. 

\textbf{Parameter setting.} There are two primary parameters in the BoF and FV models: patch size and dictionary size (or equally, the number of components of the GMM in the FV model). We tune these parameters by five-fold cross-validation on the union of training and validation sets, with the criterion of MCA. The candidate patch sizes are $9\times9$, $11\times11$, $13\times13$, $15\times15$ and $20\times20$, while the candidate dictionary sizes are $1,000$, $2,000$, $3,000$, $4,000$, $5,000$ and $10,000$. Also, the number of Gaussian components will be chosen from $64$, $128$, $256$, $512$ and $1024$ for FV. Through the cross-validation, the patch size and the dictionary size in the BoF model are selected as $15\times15$ and $10,000$. With the use of SPM, this results in a $80,000$-dimensional representation for each cell image. For the FV model, the patch size is chosen as $20\times20$ and the number of GMM components is $512$. With the use of SPM, this leads to a $262,144$-dimensional representation for each image.    

\textbf{Comparison results.} The BoF, FV and CNN models are compared on the same training and test sets. Also, both of the cases, i.e., with and without data augmentation, are investigated. To be fair, when data augmentation is used, the visual dictionary in the BoF and FV models will be built with the augmented training set. Also, to keep consistent with the setting of our deep CNN system, each test image in this case will be equally augmented and its label is predicted in the way similar to Eq.(\ref{eqn:joint-classify}), except that the probabilities are replaced by the decision values of the linear SVM classifier. 

As shown in Table \ref{Table:BoF-FV-CNN}, FV is consistently better than BoF, regardless of whether data augmentation is applied or not. This agrees well with the literature. Furthermore, both BoF and FV can well benefit from data augmentation, with an average performance increase of about $4$ percentage points. Compared with BoF and FV, CNN system shows slightly lower performance ($88.85$\% vs $89.83$\% for BoF and $91.60$\% for FV), when there is no augmentation. However, CNN outperforms both BoF and FV once data augmentation is applied. In specific, the highest MCA, $96.76$\%, is obtained by our CNN, while BoF and FV achieve only $94.23$\% and $95.73$\% respectively. Similar situation can be observed from the ACA values. These results suggest that i) when training samples are not sufficient, the high-capacity CNN is more difficult to train than the shallower, hand-designed models such as BoF and FV; and ii) by properly using data augmentation to generate more training data, the CNN can be better trained and are able to achieve better performance than the BoF and FV models. 

\begin{table*}[ht]
\centering
\resizebox{\linewidth}{!}{\begin{tabular}{c c c c c c}
\hline
\tabincell{c}{\textbf{Accuracy}\\(on test set)} & \textbf{Methods} & \tabincell{c}{No\\data augmentation} & \tabincell{c}{Augmentation by\\ a rotation angle step of $36^\circ$} & \tabincell{c}{Augmentation by\\a rotation angle step of $18^\circ$} & \tabincell{c}{Augmentation by\\ a rotation angle step of $9^\circ$} \\
\hline
\multirow{3}{*}{MCA (\%)} & BoF & 89.83 & 94.23 & 93.98 &94.14  \\ 
         & FV  & \textbf{91.60} & 95.41  & 95.73 & 95.53  \\ 
         & CNN & 88.58 & \textbf{95.99} & \textbf{96.71} & \textbf{96.76}  \\ 
\hline
\multirow{3}{*}{ACA (\%)} & BoF & 90.70 & 94.30 & 94.19 &94.38 \\ 
         & FV & \textbf{92.65} & 95.78 & 96.07 & 95.81 \\ 
         & CNN & 89.04 & \textbf{96.51} & \textbf{97.10} & \textbf{97.24}  \\                 
\hline
\end{tabular}}
\caption{Comparison of classification accuracy among the methods of BoF, FV and our deep CNN on ICPR2014 datatset}\label{Table:BoF-FV-CNN}
\end{table*}

\subsection{Experiments on the Adaptability across Datasets}
As previously mentioned, HEp-$2$ cell image classification varies with laboratory settings, the types of staining patterns involved, and the size of dataset. Such differences can be well seen from the ICPR2014 and ICPR2012 datasets. As a result, it is highly desired that a cell classification system trained with one dataset can be conveniently adapted to another one. Owning this feature not only improves the efficiency of system building, but also can take full advantages of the image data in different datasets. To demonstrate this feature for our CNN-based system, we compare the CNN purely trained on ICPR2012 dataset (called CNN-Standard in short) with the other CNN which is an adapted version of the CNN pre-trained on ICPR2014 dataset to ICPR2012 dataset (called CNN-Finetuning).   

Following previous experimental settings, CNN-Standard is trained with the $721$ training images predefined in ICPR2012 dataset. Only the green channel of each image is kept and the same preprocessing in Section~\ref{subsec:image-pre} is performed. The dropout strategy (with ratio of $0.5$) is used, because it can benefit network training and classification performance on this small dataset. CNN-Standard is trained by $100$ epochs and then used to classify the predefined test images by following Eq.(\ref{eqn:joint-classify}). 

To train CNN-Finetuning, we first select a basic CNN system learned with the ICPR2014 dataset. It is the one obtained at the $100$th epoch when the system is trained with an augmented (rotation with an angle step of $9^\circ$) training set of ICPR2014. Afterwards, this basic system is fine-tuned with the training set of ICPR2012 dataset, with or without data augmentation. All the trainable network parameters of different layers are updated during this fine-tuning process. To demonstrate the efficiency, we only fine-tune this basic system by $10$ epochs, which takes significantly less time than the $100$ epochs spent in training CNN-Standard.
\begin{figure}[h]
\begin{center}
\includegraphics[width=6.5cm]{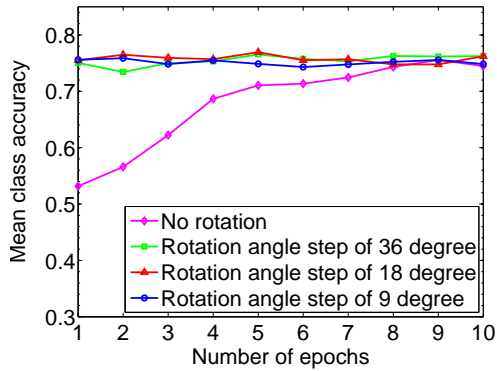}
\caption{The MCA of test set obtained by CNN-Finetuning at each of the $10$ epochs. Data augmentation with various angle steps is investigated.}
\label{fig:finetune}
\end{center}
\end{figure}

The evolution of the MCA on test set with the $10$ epochs is plotted in Fig. \ref{fig:finetune}. As shown by the line of ``No rotation'', CNN-Finetuning does not work well at the beginning. Nevertheless, it catches up quickly in a couple of epochs and reaches a satisfying performance in $10$ epochs. Furthermore, the adaption stage is significantly shortened, by applying data augmentation to the small training set of ICPR2012 to increase training samples. These results demonstrate the high efficiency of the adaptability of our CNN-based system, especially considering that there are two different classes of staining patterns across these datasets. Comparison of CNN-Standard and CNN-Finetuning is shown in Table~\ref{Table:adaption}. It is interesting to note that CNN-Finetuning consistently outperforms CNN-Standard, even though it is only fine-tuned for a few epochs. We attribute its superiority to the good initialization of the network obtained from the training process on ICPR2014 dataset. Based on the above results, we believe that our CNN-based system will be a better option for practical applications. 
\begin{table*}[ht]
\centering
\resizebox{\linewidth}{!}{\begin{tabular}{c c c c c c}
\hline
 \tabincell{c}{\textbf{Accuracy}\\(on test set)}& \textbf{Methods} &\tabincell{c}{No\\data augmentation} & \tabincell{c}{Augmentation by\\a rotation angle step of $36^\circ$} & \tabincell{c}{Augmentation by\\a rotation angle step of $18^\circ$} & \tabincell{c}{Augmentation by\\ a rotation angle step of $9^\circ$}  \\
\hline
\multirow{2}{*}{MCA (\%)} 
                                          & CNN-Standard & 63.1 & 72.4  & 72.4 & 73.2  \\ 
                                          & CNN-Finetuning & 74.5 & 76.3  & 76.2 & 74.9  \\
\hline
\multirow{2}{*}{ACA (\%)} 
                                          & CNN-Standard & 64.3 & 70.2 & 70.0 & 70.1 \\ 
                                          & CNN-Finetuning & 72.9 & 74.8  & 74.7 & 73.3 \\
\hline
\end{tabular}}
\caption{Classification accuracy of our CNN-based system on ICPR2012 dataset}\label{Table:adaption}
\end{table*}

At last, we compare our CNN-Finetuning (rotation with an angle step of $36^\circ$) with other methods reported in the literature in Table \ref{Table:compare2literature}. As seen, it outperforms the best-performing method of that contest and the CNN at the ICPR2012 contest. For that CNN, a $100 \times 100$ pixels area of the green channel centered at the largest connected component of each cell is taken via the mask and then is normalized by mapping the first and $99$th percentile values to $0$ and $1$. The architecture of that CNN is composed of two sequences of convolution, absolute value rectification and subtractive normalization, one average pooling layer, one max pooling layer and one fully connected layer\footnote{Please refer to the contest report available at \url{http://mivia.unisa.it/hep2contest/HEp2-Contest_Report.pdf} for the detailed presentation of the contest CNN.}, which is also quite different from our architecture. The better performance of our CNN may benefit from these differences as well as our effective data augmentation. Also, our CNN-Finetuning is just slightly inferior to the method in \citet{Theodorakopoulos20142367}. That method combines two kinds of hand-crafted features: the distribution of SIFT and gradient-oriented co-occurrence LBP, and a dissimilarity representation of an image is created with them. 

\begin{table}[H]
\centering
\resizebox{\linewidth}{!}{\begin{tabular}{l | c}
\hline
\textbf{Method} & \textbf{Average classification accuracy (ACA)} \\
\hline
\tabincell{l}{2012 contest\\ best-performing method \citep{Benchmarking2012}} & 68.7\% \\ 
\hline
\tabincell{l}{2012 contest CNN \citep{Benchmarking2012}} & 59.8\% \\
\hline
\citet{Nosaka20142428} & 68.5\% \\ 
\hline
\citet{Shen20142419} & 74.4\%  \\ 
\hline
\citet{Faraki20142348} &70.2\% \\ 
\hline
\citet{Larsen6802403} & 71.5\% \\ 
\hline
\citet{Theodorakopoulos20142367} & \textbf{75.1}\% \\ 
\hline
Our CNN-Finetuning & \textbf{74.8}\% \\
\hline
\end{tabular}}
\caption{Comparison with other methods on the ICPR2012 dataset}\label{Table:compare2literature}
\end{table}
In addition, it is worth mentioning that in the ICPR2014 contest \citep{ICPR2014Report}, the three methods that perform better than or comparable to our deep CNNs system ($87.10$\%, $83.64$\% and $83.33$\% vs $83.23$\% with the MCA criterion) are all built on two-stage frameworks: hand-designed feature representation and classification. The top-ranked method utilizes multi-scale and multiple types of local descriptors \citep{6973545}; the second-ranked method adopts the hand-crafted rotation invariant dense scale local descriptor \citep{6973537}; and the third method combines morphological features and different local texture features \citep{6973544}. In contrast, our CNN system generates discriminative features from raw pixels directly by utilizing class label information and jointly learns the classifier in a single architecture without learning extra dictionaries as these methods.

\subsection{Discussion on Computational Issues}
For the CNN-based classification system, training the network is the most time-consuming step in the whole pipeline. However, this process can be well accelerated by utilizing GPU programming. Also, as previously shown, an existing CNN-based system can be efficiently transferred to a new but related task via a short training process. Once the networks are trained, a test cell image only needs to go through the four networks and then is classified within $1.2$ seconds in total with Matlab implementation on a computer with $3.30$GHz Intel CPU and $16$GB RAM. 

For the BoF and FV models, building visual dictionary or the GMM is computationally intensive, especially when there are a large number of training images, e.g., due to the use of data augmentation. For example, building a dictionary of $10,000$ visual words and the GMM of $512$ components takes more than $4$ days and $2$ days in our implementation, when the training set of ICPR2014 dataset is augmented by rotation with an angle step of $9^\circ$. Also, a large dictionary in the BoF model could slow down the encoding process, e.g., around $78$ seconds per image in our experiment. Although the time for this process can be reduced in the FV model, it still takes about three seconds per image. In addition, SPM is usually needed to attain better classification performance. In this case, the dimensions of the resulting image representation are much higher than that in the CNN-based system ($80,000$ or $262,144$ vs $150$ only). 

\section{Conclusion} \label{sec:conclusion}
This paper proposes an automatic HEp-$2$ cell staining patterns classification framework with deep convolutional neural networks. We give a detailed description on various aspects of this framework and carefully discuss a number of key issues that could affect its classification performance. Extensive experimental study on two benchmark datasets demonstrates i) the advantages of our framework over the well-established image classification models on cell image classification; ii) the importance and effectiveness of data augmentation, especially when training images are not sufficient; iii) the desirable adaptability of our CNN-based system across different datasets, which makes our system attractive for practical tasks.
Much future work can be done to further improve the performance of the proposed system. In particular, a super-CNN trained with a large-scale generic image benchmark, ImageNet \citep{deng2010does}, has recently prevailed on many generic visual recognition tasks. We would like to explore the effectiveness of the features generated by this CNN for HEp-$2$ cell image and the adaption of this CNN to cell image classification. These issues will be of significance considering the substantial differences between generic images and HEp-$2$ cell images. 

\section*{References}
\bibliographystyle{model2-names}
\bibliography{Journal_Cell_Bibliography}







\end{document}